\documentclass[11pt,a4paper]{article}
\usepackage[hyperref]{emnlp2020}
\usepackage{times}
\usepackage{latexsym}
\usepackage{xcolor,colortbl}
\usepackage{amsmath}
\usepackage{graphicx}
\usepackage{booktabs}
\usepackage{caption}
\usepackage{subcaption}

\usepackage{color}
\usepackage{multicol}
\usepackage{url}
\usepackage{multirow}
\usepackage{amssymb}
\usepackage{tabularx}
\usepackage{booktabs}

\usepackage{microtype}

\title{Social Commonsense Reasoning with Multi-Head Knowledge Attention}

\aclfinalcopy
\author{Debjit Paul \\
  Research Training Group AIPHES \\
  Institute for Computational Linguistics\\
  Heidelberg University \\
  {\tt paul@cl.uni-heidelberg.de} \\\And
  Anette Frank \\
  Research Training Group AIPHES \\
  Institute for Computational Linguistics \\
  Heidelberg University\\
  {\tt frank@cl.uni-heidelberg.de} \\}

\date{}

\begin{document}
\maketitle
\begin{abstract}

Social Commonsense Reasoning requires understanding of text, knowledge about social events and their pragmatic implications, as well as commonsense reasoning skills. In this work we propose a novel \textit{multi-head knowledge attention model} that encodes semi-structured commonsense inference rules and learns to incorporate them in a transformer-based reasoning cell. We assess the model's performance on two tasks that require different reasoning skills: \textit{Abductive Natural Language In\-fer\-ence} and \textit{Counterfactual Invariance Prediction} as a new  task.
We show that our proposed model improves performance over strong state-of-the-art models (i.e., RoBERTa) across both reasoning tasks. Notably we are, to the best of our knowledge, the first to demonstrate that a model that learns to perform \textit{counterfactual reasoning} helps predicting the best explanation in an \textit{abductive reasoning} task.
We validate the robustness of the model's reasoning capabilities by perturbing the knowledge and provide 
qualitative analysis on the model's knowledge incorporation capabilities. 

\end{abstract}

\newcolumntype{C}{>{\centering\arraybackslash}p{2.5cm}}
\renewcommand*{\arraystretch}{1.2}
\begin{table*}[!tbp]
      \scalebox{0.8}{
      \centering
      \begin{tabular}{@{}p{10mm}|p{155mm}|>{\arraybackslash}p{20mm}}
          \hline
            \rowcolor{gray!15}
            \textbf{Task} &\textbf{Context} & \textbf{Answer} \\ \hline
            {$\alpha$NLI} & {$O_1$:} \textit{Dotty was being very grumpy.}& \\
            & \hspace*{1cm} $H_1$: Dotty ate something bad.& \textbf{$H_1$} or {$\mathbf{H_2}$} \\
            & \hspace*{1cm} $H_2$: Dotty call some close friends to chat. & \\
            &$O_2$: \textit{She felt much better afterwards.} &\\ \hline
            {CIP} &  \textit{$s_1$: Bob had to get to work in the morning.} & \\
            & \hspace*{1cm} \textit{$s_2$: His car battery was struggling to start the car.} \textit{$s_3$: He called his neighbor for a jump start.} & \\
            {} & \hspace*{1cm}  \textit{$s_2'$: His car won't start.} \hspace*{4.1cm}\textit{$s_3$: He called his neighbor for a jump start.} & \textbf{[Yes]} or [No] \\
            {}& \textit{$s_1$: Bill and Teddy were at the bar together.}&\\ &\hspace*{1cm} \textit{$s_2$: Bill noticed a pretty girl.} \hspace*{9mm} \textit{$s_3$: He went up to her to flirt.} &  \\
            &\hspace*{1cm} \textit{$s_2'$: Bill noticed his mom was there.}  \textit{$s_3$: He went up to her to flirt.} &  [Yes] or \textbf{[No]}        \\
            \hline
      \end{tabular}}
      \caption{Examples from each dataset used in this work. The correct choice in each example is given in bold text.} 
      \label{tab:example}
\end{table*}

\section{Introduction}
\begin{figure}[t]
  \centering
    \includegraphics[scale=1.0,height=5cm, width=0.36\paperwidth]{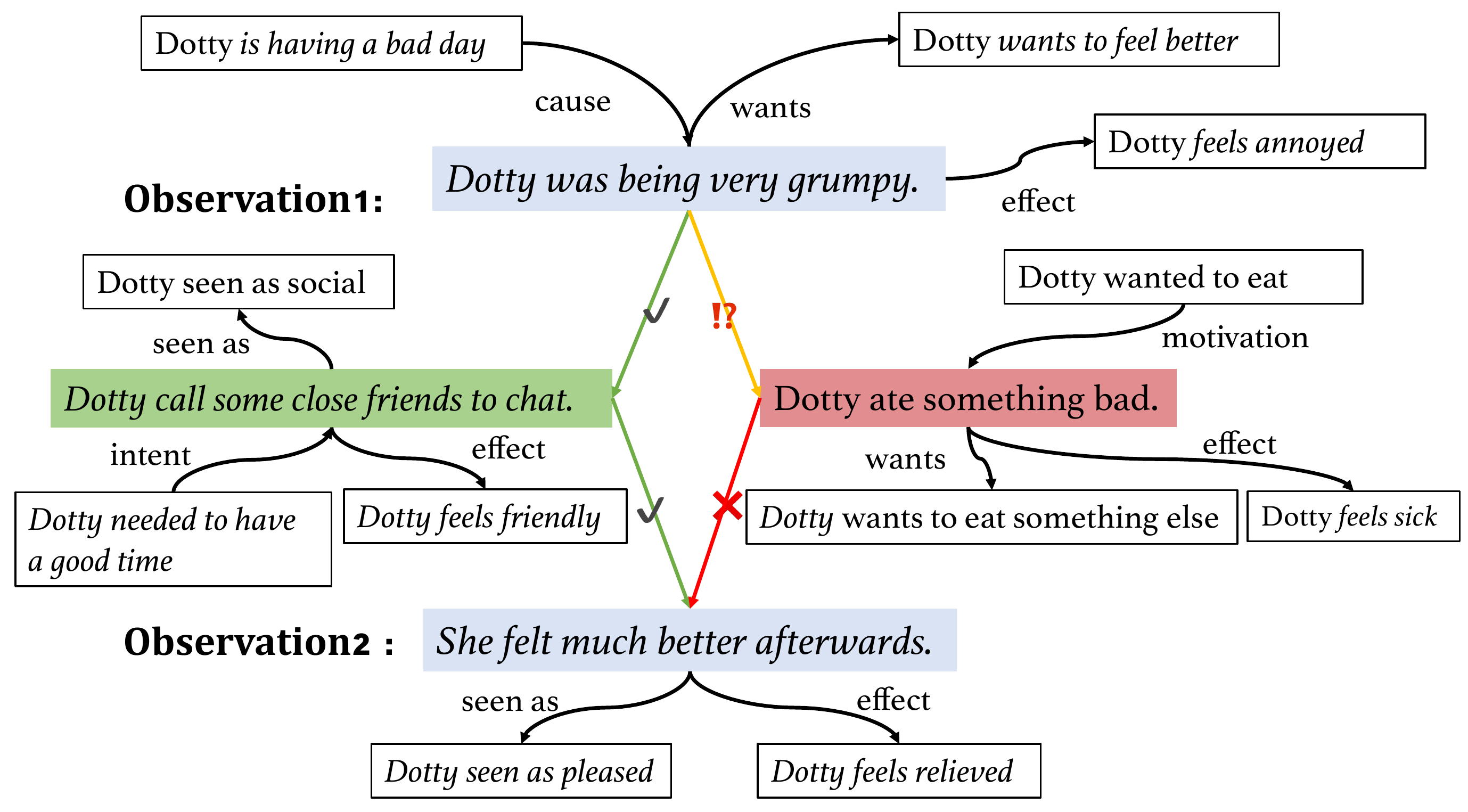}
    \caption{Motivational example: The top and bottom blue boxes show two observations. The green and red box contain a plausible and an implausible hypothesis, respectively. A green line denotes that an event is likely to follow, the yellow line that an event is somewhat unlikely to follow, the red line something unlikely.}
    \label{fig:example_motivation}
\end{figure}

{Humans are able to understand natural language text about everyday situations effortlessly, by relying on 
commonsense knowledge and making inferences.}
For example in Figure \ref{fig:example_motivation}, given two observations:
\textit{Dotty was being very grumpy} and 
\textit{She felt much better afterwards} 
-- we can come up with a
plausible explanation about what could have provoked the change in Dotty’s emotion. We can also construct alternative hypotheses that will not change Dotty’s emotion.
In order to judge the plausibility of such explanations, we need to have information about
mental states and social norms, i.e.,  a form of commonsense knowledge. Such information
includes that \textit{calling a close friend}, in general, makes \textit{people feel happy}.
This kind of inference 
goes beyond the broadly studied textual entailment task \cite{bowman2015large} in that i) it requires a specific form of knowledge, namely knowledge about mental states (\textit{intent, motivation}), social norms (\textit{cause or effect of an event}) and behaviour (\textit{emotional reactions}),  and ii) the awareness that inferences we can draw on their basis 
must often be viewed as
plausible explanations, and hence can be defeasible, rather than being strict inferences.

In this paper, we investigate social commonsense reasoning in 
narrative contexts. Specifically, we address two different reasoning tasks: language-based abductive reasoning, and counterfactual invariance prediction. We introduce the Counterfactual Invariance Prediction task (CIP), which tests the capability of models to predict whether under the assumption of a counterfactual event, a factual event remains invariant or not in a narrative context. Figure \ref{fig:example_motivation} illustrates an example: Given a narrative context -- \textit{``Dotty was being very grumpy'' (premise), ``Dotty called some close friends to chat'' (hypothesis), ``She felt much better afterwards.''(conclusion)} -- will a counterfactual assumption
(alternative hypothesis), e.g., \textit{``Dotty ate something bad''}, still lead to same conclusion? 

While there has been positive impact of trans\-former-based pretrained language models (LMs) \cite{devlin-etal-2019-bert, Liu2019RoBERTaAR} on several downstream NLU tasks including commonsense reasoning, there is still a performance gap between machines and humans, especially when the task 
involves implicit knowledge \cite{Talmor2018CommonsenseQAAQ}.

There are two important bottlenecks: (i) obtaining relevant commonsense knowledge and (ii) effectively incorporating it into state-of-the-art neural models to improve their reasoning capabilities.
 
In current research, the standard approach to address the first bottleneck is 
to extract knowledge tuples or paths from large structured knowledge graphs (KGs) (e.g. ConceptNet, \citet{speer2017conceptnet}) using graph-based methods 
\cite{bauer2018commonsense, Paul2019RankingAS, lin2019kagnet}. However, in this work, instead of retrieving and selecting knowledge from a static KG, 
we \textit{dynamically generate contextually relevant knowledge} using COMET (based on GPT-2) \cite{bosselut-etal-2019-comet}.
To address the second bottleneck, we build on the 
hypothesis
that models performing such reasoning tasks need to 
consider multiple knowledge rules jointly (see Fig.\ \ref{fig:example_motivation}). 
Hence, we introduce a novel \textit{multi-head knowledge attention model} which learns to focus on multiple pieces of knowledge at the same time, and is able to refine the input representation in a recursive manner, 
to improve the reasoning capabilities.

An
important aspect of 
using specified knowledge rules
is 
a gain in
interpretability. 
In this work, we perturb the pieces of knowledge available to the model to demonstrate
its robustness, and we provide 
qualitative analysis to offer deeper insight into the model's capabilities.

Our contributions are: i) We propose a new \textit{multi-head knowledge attention model} that uses structured knowledge rules to emulate reasoning.
ii) We compare our model with several state-of-the-art neural architectures for QA tasks and show that it performs better
on \textit{two 
types of reasoning tasks}.
iii) We specifically compare our novel \textit{knowledge integration technique} to prior integration methods
and show it performs better on the abductive reasoning task (+2
percentage points).
iv) We introduce a \textit{novel counterfactual invariance prediction} (CIP) task, and show a \textit{correlation} between \textit{abduction} and \textit{counterfactual reasoning} 
in a narrative context. 
v) To \textit{analyze
the reasoning capabilities} of our model we investigate 
a) 
how it performs without fine-tuning on a pre-trained 
model, b) how robustly it behaves when confronted with
perturbations and noise in the knowledge and c) offer qualitative analysis of the reasoning module.

Our code is made publicly available.\footnote{\url{https://github.com/Heidelberg-NLP/MHKA}} 


\if false
\begin{table*}[!tbp]
      \scalebox{0.8}{
      \centering
      \begin{tabular}{@{}p{28mm}|p{100mm}|p{60mm}}
          \hline
            \rowcolor{gray!15}
            \textbf{Dataset} &\textbf{Context} & \textbf{Answer} \\ \hline
            {Abductive Natural Language Inference} & {Observation1:} \textit{Dotty was being very grumpy.} \textit{Observation2:} \textit{She felt much better afterwards.} & Hypothesis1: Dotty ate something bad. \\ {}&{}&\textbf{Hypothesis2: Dotty call some close friends to chat.} \\ \hline
            {Counterfactual Invariance Prediction} &  \textit{$s_1$: Bob had to get to work in the morning.}
            \textit{$s_2$: His car battery was struggling to start the car.} 
            \textit{$s_3$: He called his neighbor for a jump start.} & \textbf{[Yes]} or [No] \\
            {} & Alternatively: \textit{$s_1$: Bob had to get to work in the morning.} \textit{$s_2'$: His car won't start.}  \\[3mm]
            {}& \textit{$s_1$: Bill and Teddy were at the bar together.} \textit{$s_2$: Bill noticed a pretty girl.}
            \textit{$s_3$: He went up to her to flirt.} & [Yes] or \textbf{[No]} \\
            {} & Alternatively: \textit{$s_1$: Bill and Teddy were at the bar together.}  \textit{$s_2'$: Bill noticed his mom was there.} 
            \\
            \hline
      \end{tabular}}
      \caption{Examples from each dataset used in this work. The correct choice in each example is given in bold text.} 
      \label{tab:example}
\end{table*}
\fi 

\section{Social Commonsense Reasoning Tasks}
\label{task}
We address two social commonsense reasoning tasks that require different reasoning skills. They are exemplified in Table \ref{tab:example} and detailed below. \\ 
\textbf{Abdutive Natural Language Inference ($\alpha$NLI)} \citet{bhagavatula2019abductive} created 
a dataset that tests a model's ability
to choose the best explanation for
an incomplete set of observations. 
Abduction is a backward reasoning task.  Given a pair of observations $O_1$ and $O_2$, the $\alpha$NLI task is to select the most plausible explanation (hypothesis)
$H_1$ or $H_2$.\\
\textbf{Counterfactual Invariance Prediction (CIP)} Counterfactual Reasoning (CR) is the mental ability to construct alternatives (i.e., counterfactual assumptions) to past events and to reason about their (hypothetical) implications \cite{epstude2008functional, roese2009psychology}. 
One of the key challenges of CR is judging \textit{causal invariance}, i.e., deciding whether a given
factual event is invariant under counterfactual assumptions, or whether it is not
\cite{peters2016causal, qin-counterfactual}. 

In this work,
we define a new \textit{Counterfactual Invariance Prediction (CIP)} task that 
tests the capability of models 
to predict whether 
under the assumption of
a counterfactual event, a (later) factual event remains invariant or not 
in a narrative context (cf.\ Table \ref{tab:example}). 
This task requires deeper understanding of causal narrative chains and reasoning in forward direction. \citet{qin-counterfactual} proposed a dataset to encourage models to learn to rewrite stories with counterfactual reasoning. We automatically collect counterfactual invariance examples along with non-invariant examples from their dataset
to create
a balanced dataset for our proposed CIP task. 

The formal setup is: given the first three consecutive sentences from a narrative story ${s_1}$ (premise), ${s_2}$ (initial context), ${s_3}$ (factual event) and an additional 
sentence $s'_2$ that is counterfactual to the initial 
context ${s_2}$, the task is to predict whether ${s_3}$ is 
invariant given $s_1, s'_2$ or not. 
The train/dev/test data (cf.\ Table \ref{tab:data_stat}) are balanced with
an equal number
of \textit{Yes/No} answers, hence the random baseline 
is 50\%.
To compute human performance, we gave 100 instances from the test set to expert evaluators. Human accuracy on the CIP task is at
\textit{84.8\%}.\footnote{More details about the data are given in the
\textit{Supplement}.
} 

\begin{figure}[t]
  \centering
    \includegraphics[scale=1.0,height=3cm, width=0.35\paperwidth]{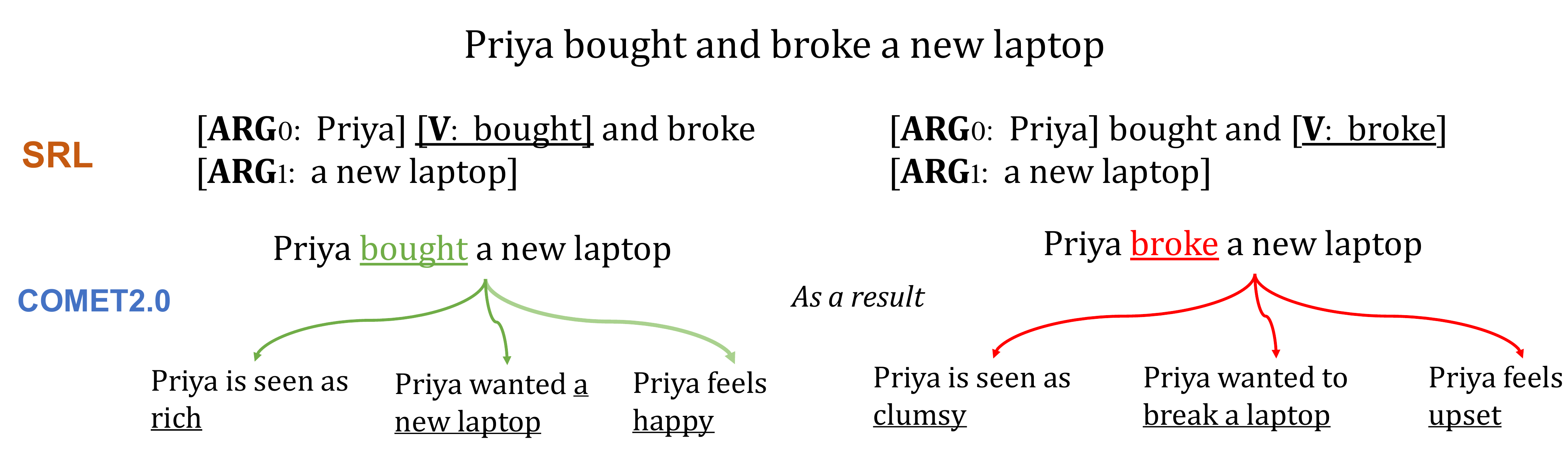}
    \caption{Depicting the steps to extract commonsense knowledge about social events.} 
    \label{fig:example_srl_comet}
\end{figure} 
\section{Semantic \& Commonsense Knowledge} \label{sec:extraction}

This section details
the steps we follow to generate social commonsense knowledge about events mentioned in a narrative. See  Figure \ref{fig:example_srl_comet} for illustration. 


Understanding a narrative text requires the ability to identify events and to reason about their 
causal effects.
Beyond causal relations, they require the understanding of narrative relations, as in narrative chains or schemata \citep{chambers-jurafsky-2008-unsupervised}. This is knowledge about characteristic script-like event sequences where semantic roles of consecutive events are referentially bound to roles of preceding events. While \citet{chambers-jurafsky-2008-unsupervised} focused on the induction of schemata using corpus statistics, we will combine detected events with deeper commonsense knowledge. 

In a first step we apply SRL to extract the basic structure \textit{``who did what to whom, when and where''} from each sentence in the context, using 
state-of-the-art SRL \cite{shi2019simple}.
In a second step, we use commonsense transformer (COMET2.0,\footnote{COMET2.0 uses GPT-2 as pretrained model.} \citet{bosselut-etal-2019-comet}) to extract social commonsense knowledge about the extracted events.
COMET2.0 is trained on the ATOMIC \cite{sapatomic} inferential knowledge resource which consists of 877K everyday events, each characterized by nine relation types (\textit{xIntent, xNeed, xReact}, etc.) which we call \textit{dimensions}. These dimensions connect the event in question with manifold properties, emotions, as well as other states or events. 

In the last processing step we generate, for each event in each sentence from our datasets, all dimensions defined for it using COMET2.0. For example,  for:
\textit{Dotty ate something bad} we generate (among others)\footnote{More examples are given in the \textit{Supplement}.} the tuple: $\langle$\textit{PersonX, xReact, sick}$\rangle$ and derive $\langle$\textit{Dotty, feels, sick}$\rangle$ by substituting \textit{PersonX} with the logical subject, the filler of the role \textit{ARG0}.


\section{A Multi-Head Knowledge Attention (MHKA) Model for Social Reasoning} \label{sec:model}
In this section we introduce
the 
MHKA model  and discuss some key differences in how MHKA works for the two different Social Commonsense Reasoning tasks. 
For a model overview see Figure \ref{fig:model}.
\subsection{Model Architecture}
MHKA consists of 3 modules: (a) the \textit{Context En\-co\-ding Layer} consists of a pre-trained LM, (b) the \textit{Knowledge Encoding Layer} consists of stacked transformer blocks, (c) the \textit{Reasoning Cell} consists of transformer blocks with \textit{multi-head attention} that allows the model to jointly attend to the input representation and the encoded knowledge. The input format for each task is depicted in Table \ref{tab:format}. 

\textbf{(a) Context Encoding Layer:} 
For each task, we concatenate the inputs as a sequence of tokens $x_n$ = ($x_{n_{1}}$, .. $x_{n_{m}}$), and compute contextualized representations with a pre-trained LM. We obtain $n$ different representations for $n$ input options  i.e., $h_{x_{n}} = encode(x_n) = (h_{n_{1}}, .. , h_{n_{m}})$, where 
for $\alpha$NLI $n$=2 and for CIP $n$=1. As pre-trained LMs we consider (i) BERT \cite{devlin-etal-2019-bert} and 
(ii) RoBERTa \cite{Liu2019RoBERTaAR}.

\begin{figure}[t]
  \centering
    \includegraphics[scale=1.0,height=7cm,width=\linewidth]{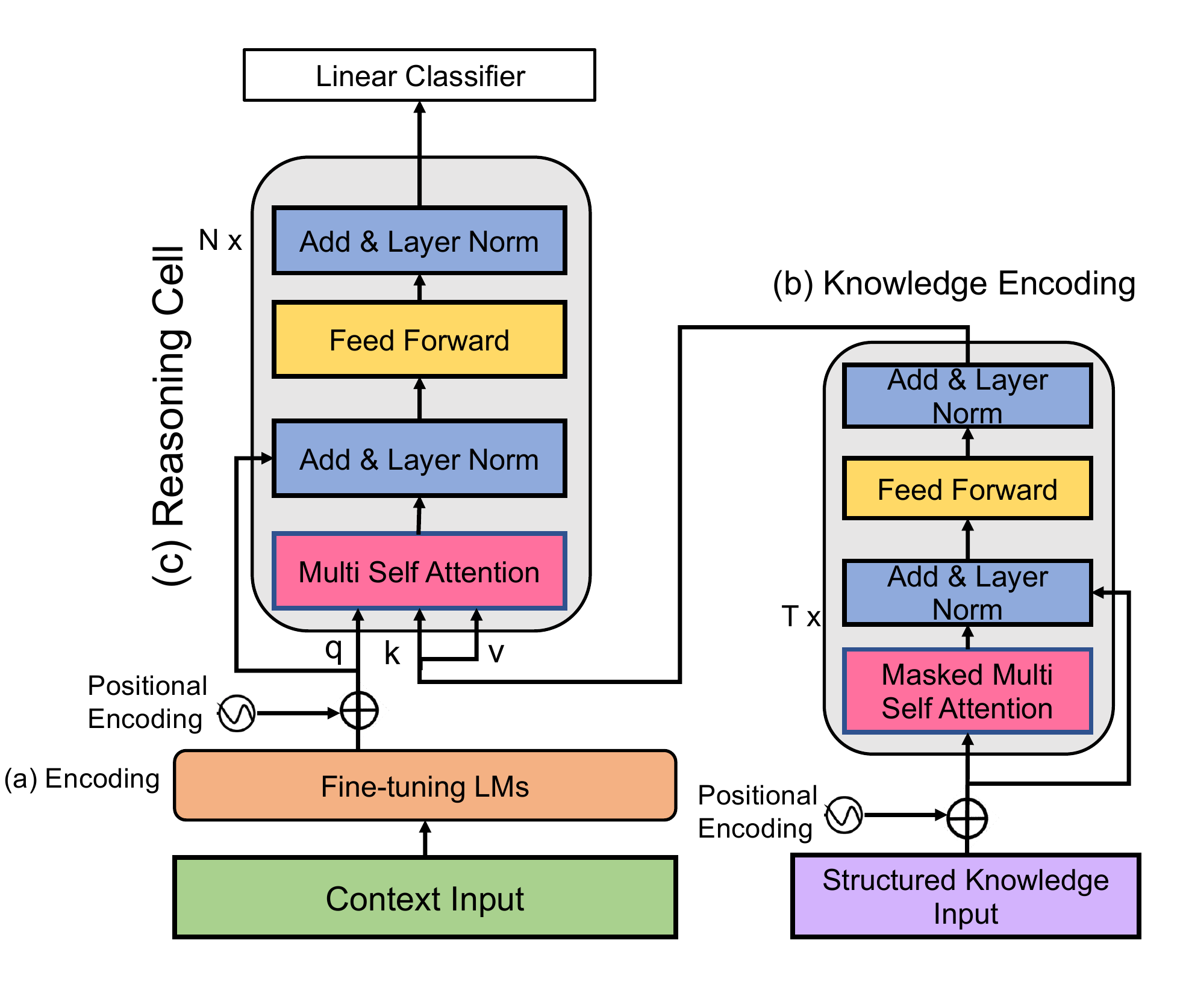}
    \caption{Overview of our Multi-Headed Knowledge Attention Model. It consist of three components (a) the \textit{Context En\-co\-ding Layer} (b) the \textit{Knowledge Encoding Layer}, and (c) the \textit{Reasoning Cell}.} 
    \label{fig:model}
\end{figure}

\textbf{(b) Knowledge Encoding Layer:} As depicted in Figure \ref{fig:model}, the knowledge encoding layer is a Transformer-Block 
\cite{j.2018generating, alt-etal-2019-fine} as typically used in the decoder part of the trans\-for\-mer model of \citet{vaswani2017attention}. 
The core idea is that the model repeatedly encodes the given knowledge input
over multiple layers (i.e., Transformer blocks), where each layer consists of masked multi-head self-attention followed by layer normalization and a feed-forward operation. Similar to the context input format, we concatenate the knowledge inputs as a sequence of tokens $k_n$ = ($k_{n{_1}}, .. , k_{n{_w}}$), where $k_n$ is the knowledge used  for input option $x_n$. 
In order to obtain the hidden knowledge representation we do the following: 
\begin{equation}
\begin{aligned}
  h_{k_{n}^{0}} = k_{n} W_{ke} + W_{kp} ,\\
  h_{k_{n}^{l}} = tb(h_{k_{n}^{l-{1}}}), \forall l \in [1,L]
\end{aligned}
\end{equation}
where $W_{ke}$ is the token embedding matrix, $W_{kp}$ the position embedding matrix, $tb$  the transformer block, and $L$  the number of transformer blocks.
\begin{table}[t!]
\small
\centering
{
\begin{tabular}{@{}l@{~}c@{~}l@{}}
\toprule
{\bf Task} & {\bf Input Format} & {Output}\\
\midrule
\textbf{$\alpha$NLI} & [CLS] $O_1$ $H_i$ [SEP] $O_2$ [SEP] & {$H_1$ or $H_2$}\\
\textbf{CIP} & [CLS] {$s_1$ $s_2$ $s_3$ [SEP] $s_1$ $s'_2$ $s_3$} [SEP] & {YES or NO}\\
\bottomrule
\end{tabular}}
\caption{Different input and output formats: [CLS] is a special token used for classification, [SEP] a delimiter.}\label{tab:format}
\end{table}

\textbf{(c) Reasoning Cell:} The main 
intuition behind the reasoning cell is that given the context representation, the model should learn to emulate reasoning over the input using the knowledge representation obtained from the knowledge encoder. 
The Reasoning Cell is another transformer block, where the model repeatedly performs multi-head attention over
the context 
and knowledge representations, and thus can iteratively refine the 
context representation. 
This capability is crucial for allow\-ing the model to emulate complex reasoning steps through composition of various knowledge pieces.
The multi-head attention function has three inputs: a query $Q$ (context re\-pre\-sen\-ta\-tion), key $K$ and
value $V$ (both knowledge re\-pre\-sen\-ta\-tion). It relies on scaled dot-product attention  
\begin{equation}
\begin{aligned}
   Q =  h_{x_{n}} + W_{xp}\\
   a_{xk_{n}} = softmax(\frac{QK^T}{\sqrt{d_z}})V
\end{aligned}   
\end{equation}
where $K$ = $V$ = $h_{k_{n}}$, $d_z$ the dimensionality of the input vectors
representing the key and value, and $W_{xp}$ is the position embedding.
We project the output representations from the reasoning cell into logit ($s$) of size $n$ (the number of output values) using a linear classifier. Finally, we compute the scores $y = max(s_i)$ where, $i = 1, .., n$. For CIP, where $n$ = 1, we treat a logit score  $>0$ as predicting yes, otherwise the answer is no. 

\subsection{Applying the MHKA model to advanced 
Social Commonsense Reasoning Tasks}\label{sec:sec_mhka}
There are some key differences in how MHKA solves the two reasoning
tasks: 

\if false
(a) In \textbf{SocialIQA*},  the model needs to perform inference about social interactions in order to answer the questions. 
Figure \ref{fig:role_mhka} shows an example where for some given question  the model is tasked to predict  \textit{Riley's intentions} (Q2) while for  another (Q1) it needs to predict 
an \textit{attribute of Riley}. 
In this task, the model reads the encoded input: 
$C$(ontext), $Q$(uestion) and candidate $A$(nswer) $i$ and predicts the correct $A_i$ (cf.\ Table \ref{tab:format}).
The Reasoning cell in MHKA 
selects and combines
the relevant knowledge rules
that support inferring 
(or choosing) the most plausible
answer. We find that in SocialIQA*, 
for most questions, predicting the correct answer requires a single knowledge rule.\\
\begin{figure}[t]
  \centering
    \includegraphics[scale=1.0,height=3.3cm, width=0.38\paperwidth]{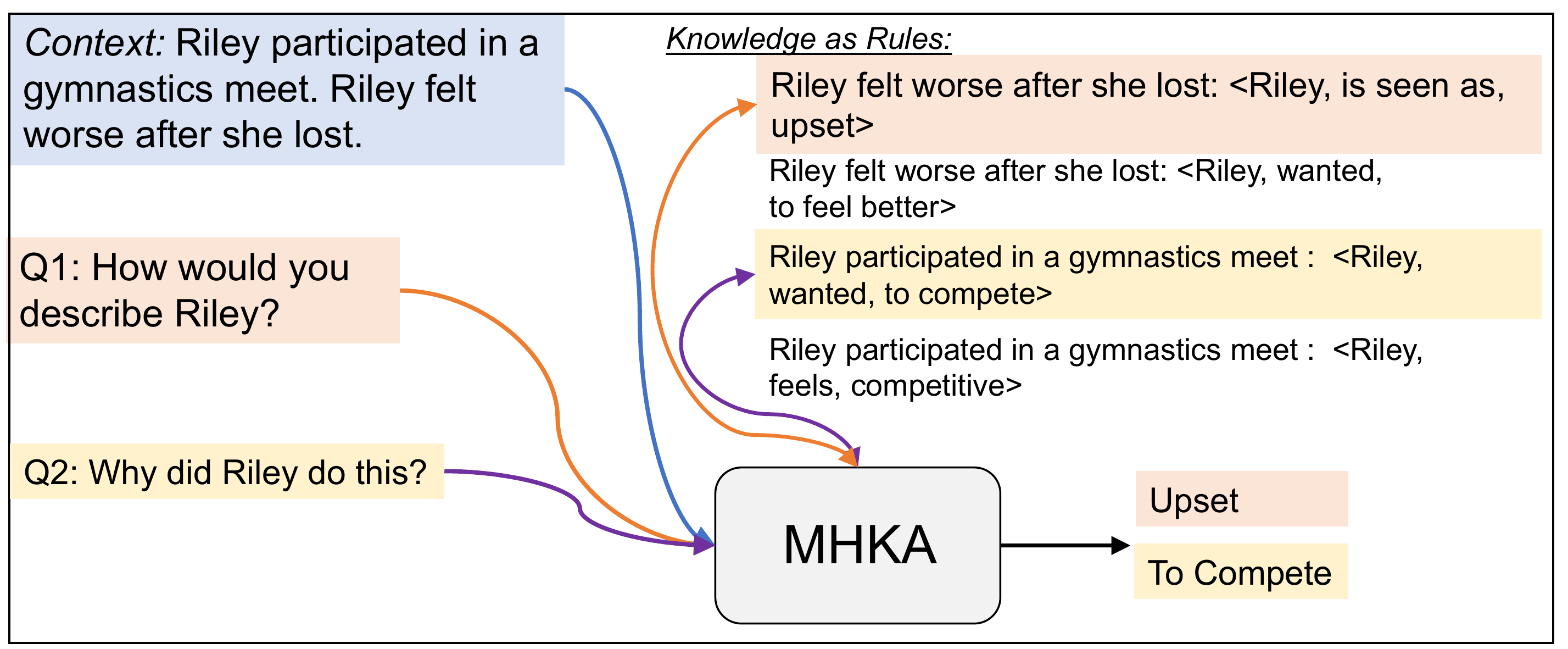}
    \caption{Depicting the mechanism of MHKA}
    \label{fig:role_mhka}
\end{figure}
\fi 

(a) In the \textbf{abductive $\alpha$NLI reasoning task}, the model must predict -- given incomplete observations $O_1$ and $O_2$ -- which of two hypotheses $H_i$ is more plausible.
For example: $O_1$: \textit{Daniel wanted to buy a toy plane, but he didn't have any money}; $O_2$: \textit{He bought his toy plane, and kept working so he could buy another}; correct $H_i$: \textit{He opened a lemonade stand}. 
Here, the model needs to link $O_2$ back to $O_1$ using social inference knowledge relating to the $H_i$ that best supports one of the sequences: $O_1, H_i, O_2$.
In this case, the model obtains the (encoded) input: $O_1$, $H_i$, $O_2$, and is tasked to predict the correct $H_i$, using available  knowledge rules.\footnote{Relevant knowledge from COMET2.0 here includes:  
[$O_1$: Daniel wanted to have money] $\,\to\,$ [$H_i$: Daniel wanted to make money, \textbf{Daniel then makes money}] $\,\to\,$ [$O_2$: Daniel needed to have money]. Clearly, $H_i$ is supported by $H_1$: \textit{He opened a lemonade stand}. So we can judge that the selected knowledge (partially) supports $H_1$.} 

(b) For \textbf{Counterfactual Invariance Prediction, CIP}, the model needs to decide whether for
given a context $C_{s_1,s_2,s_3}$, under the assumption of a counterfactual $s_{2}'$, the given
$s_3$ remains invariant or not. I.e., given: \textit{Dotty was 
grumpy. Dotty called 
close friends to chat. She felt 
better afterwards.} and the counterfactual $s_2'$: 
\textit{Dotty ate something bad} -- can it still be true that \textit{Dolly felt better afterwards}? 
Here our model gets as input the factual ($s_2$) and a counterfactual $(s'_2$) context: $s_1$, $s_2$, $s_3$ [SEP], $s_1$, $s_2'$, $s_3$ (cf.\ Table \ref{tab:format}) and is tasked to predict whether or not $s_3$ remains true under the assumption $s_2'$.
Again, the model needs to identify relevant knowledge to substantiate whether $s_3$ prevails given $s_1$ and $s_2'$.

\textbf{Abduction \textit{meets} Counterfactual Reasoning} 
Clearly, when learning to judge whether $s_3$ holds true given both a factual ($s_1, s_2$) and  counterfactual ($s_1, s_2'$) context, the CIP model learns how different events can or cannot lead to the very same factual event in a hypothetical reasoning task.
Our intuition is that such a model effectively also acquires knowledge about what kinds of events can provide  \textit{evidence} for a given event, as is needed to perform \textit{abduction}.
Hence, we hypothesize that a model that has learned to understand and reason about counterfactual situations 
can also 
support 
abductive reasoning (i.e., finding the best explanation for an event). 
In our experiments, we test this hypothesis, and evaluate the performance of a model on the $\alpha$NLI task,
that we first train
on CIP and then
finetune it on the abductive inference task. 

\if false
\textbf{Abductive Reasoning \textit{meets} Counterfactual Reasoning} 
We further explore the relationship between abductive reasoning and counterfactual reasoning. 
For CIP task the MHKA model during the encoding looks at both the initial context and counterfactual at the same time and learns how different events can lead to same factual events or not (can be seen as a forward inference task). Our intuition is that such model should posses knowledge about things that lead to what has occurred (knowledge about \textit{cause}). In this work, we hypothesize that a model that learns to understand and reason about counterfactual situations (\textit{i.e., understanding how contrary to what actually happened, might or might not have resulted in different factual events}) will help the model to perform abduction (i.e., find the best explanation) better. In our experiments below, we will test this hypothesis, and evaluate the performance of a model in an \textit{abductive inference task} while first training it on a \textit{counterfactual} and transfer the learning to abductive inference task in a narrative setting. 
\fi 
\if false
\begin{figure}[t]
  \centering
    \includegraphics[scale=1.0,height=2.5cm, width=0.38\paperwidth]{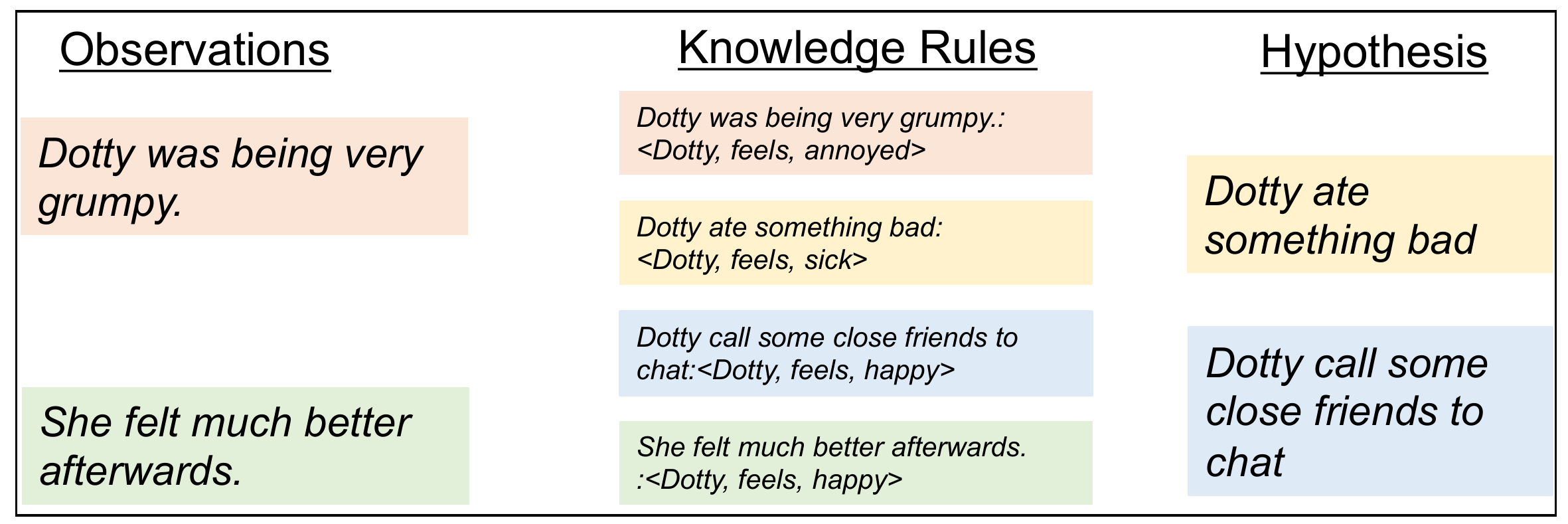}
    \caption{Depicting the change in a person's feeling in a narrative chain \af{[one needs to understand that red box is O1 and blue box is O2; also the semantics of colors is not clear: why is Dotty-feels-happy two times given, once green and once blue?]}}
    \label{fig:narrative_change}
\end{figure}

In the field of Cognitive Science,  Philosophy of language, and Philosophy of mind there has been wide study about comparative similarity and counterfactual plausibility. According to David Lewis \cite{lewis1973counterfactuals, lewis1973counterfactual, lewis1979counterfactual}, we humans evaluate the plausibility that a counterfactual event might have occurred by comparing an imagined possible world in which the counterfactual statement is true against the actual world where the counterfactual statement is false (cf.\\citet{stanley2017counterfactual}). 

Accordingly, given there are two counterfactual situations ($e_1,e_2$) which are true in two possible worlds $W_1, W_2$ respectively, now if one world (lets say, $W_1$) is comparatively more close to the real world than the other world ($W_2$) then, the first situation ($e_1$) is more plausible than the other one ($e_2$) \cite{sep-causation-counterfactual}. For example, imagine two possible situations : ($e_1$) \textit{``Dotty was being very grumpy. Dotty ate something bad. She felt much better afterwards.''} is true in $W_1$, ($e_2$) \textit{``Dotty was being very grumpy. Dotty called some close friends to chat. She felt much better afterwards.''} is true in $W_2$. Now, comparing both worlds to the real world, the one that is found to be closer, is judged more plausible i.e., $W_2$ in this example. Therefore, we hypothesize that a model that learns to understand and reason about counterfactual situations (\textit{i.e., identifying counterfactual situations that could have occurred}) will help the model to perform abduction (i.e., find the best explanation) better. 
\af{In our experiments below, we will test this hypothesis, and evaluate}
the performance of a model \af{in} an  \textit{abductive inference task} \af{-- while training it} 
on a \textit{counterfactual (\textit{plausiblity}) reasoning task} in a narrative \af{(setting?)} context \af{(cf.\ Table \ref{tab:mixed})}.
\af{[OK, but I have two problems with how you describe counterfactuals in this section: 1. you mix up how counterfactuals are discussed in two different scientific contexts: cognitive science and philosophy of language; 2. yes, similarity (or plausibility) of worlds is important here, but one should not forget that counterfactuals are about assumption (restrictor, p) and conclusion (q). But you only talk about the similarity of worlds where p is true - while most relevant is whether, under condition that p is true in some world, the real question is whether q holds too, IN EXACTLY these worlds that are maximally similar to the actual world, BUT where p is true (and not not-p). This is missing, and it was also missing in lines 229ff (see my notes). \\
So it is not that you have to compare e1 and e2 by themselves, and choose e2 because it is more similar: the question is: if e2, the more similar world is true, does then q hold in exactly this world W2 (and if it is a universal quantification, does q hold in all those worlds where e2 is true and that are maximally similar to the factual world, too?]}
\debjit{I see your point}
\fi 


\if false
\begin{figure}[t]
  \centering
    \includegraphics[scale=1.0,height=3cm]{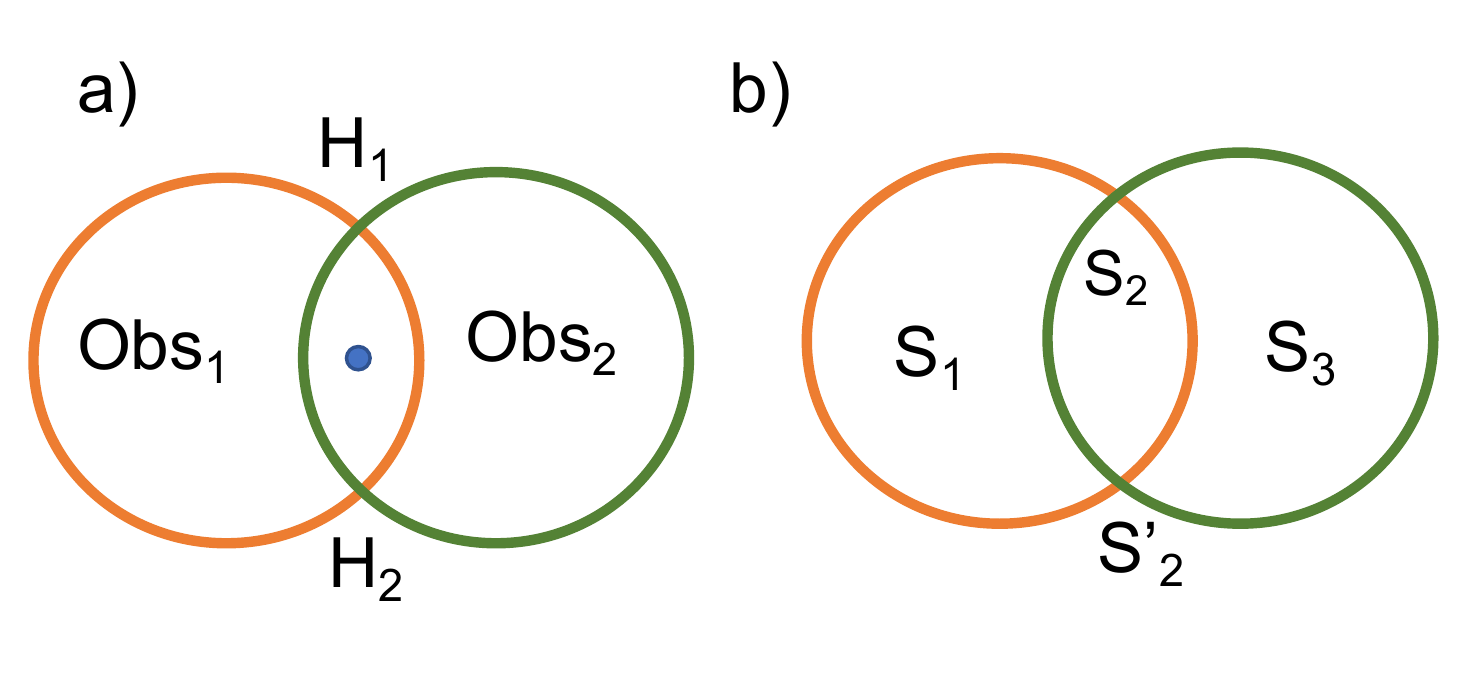}
    \caption{Abduction meets Counterfactual}
    \label{fig:amc}
\end{figure}
\fi

\section{Experiments}

\begin{table}[t!]
\small
\centering
{
\begin{tabular}{@{}llll@{}}
\toprule
{\bf Task} & {\bf Train}& {\bf Dev} & {\bf Test}\\
\midrule
$\alpha$NLI & {169654}& {1532} & {3059} \\
CIP & {12700}& {1008} & {1184}\\
\bottomrule
\end{tabular}}
\caption{Dataset Statistics:
nb.\ of instances.}\label{tab:data_stat}
\end{table} 

\paragraph{Tasks and Settings.} We apply our model to the two
social reasoning tasks introduced in \S \ref{task}. We train models for each task using the input settings stated in Table \ref{tab:format}. Data statistics is given in Table \ref{tab:data_stat}. We extract,  for each event in each input sentence, social commonsense reasoning knowledge from COMET2.0, as detailed in \S \ref{sec:extraction}. For the extraction process we use SRL as implemented in AllenNLP \cite{gardner-etal-2018-allennlp}.


\paragraph{Hyperparameter Details.}
In all models the Reasoning Cell and the Knowledge Encoder are both instantiated by  a Transformer with $4$ attention heads and depth=$4$. 
For each task, we 
select the hyperparameters that yield best performance on the dev set.
Specifically, we perform a grid search over the hyperparameter settings with a learning rate in \{$1e$-$5$, $2e$-$5$, $5e$-$6$\}, a batch size in $\{4, 8\}$, and a number of epochs in $\{3, 5, 10\}$.  
Training is performed using cross-entropy loss. For evaluation, we measure accuracy.
We report performance on the test sets 
by averaging results along with the variance obtained for $5$ different seeds. 
See \textit{Supplement} for details.

\paragraph{Baselines.} 

We compare our model to the following baselines:\\
(a) \textit{OpenAI-GPT} \cite{radford2018improving} is a multi-layer Transformer-Decoder based language model, trained with an objective to predict the next word. \\
(b) \textit{Transformer Encoder} Model has the same architecture\footnote{12-layer, 768-hidden, 12-heads} as OpenAI-GPT without pre-training on large amounts of text. \\
(c) \textit{BERT} \cite{devlin-etal-2019-bert} is a LM trained with a masked-language modeling (MLM) and next sentence pre\-dic\-tion objective, i.e.,\ it is trained to predict words that are masked from the input. \\
(d) \textit{RoBERTa} \cite{Liu2019RoBERTaAR} has the same architecture as BERT, yet without next-sentence prediction objective. \textit{RoBERTa-B(ase)} and \textit{-L(arge)} were trained on more data and optimized carefully.\\
(e) \textit{McQueen} \cite{mitra2019exploring} proposed ways to infuse unstructured knowledge into pretrained language model (RoBERTa) to address the $\alpha$NLI task. \citet{mitra2019exploring} used original \textit{ROCStories} Corpus \cite{mostafazadeh-etal-2016-corpus} and Story Cloze Test
that were used in creating $\alpha$NLI dataset. \\
(f) $L2R^2$ (Learning to Rank for Reasoning) \cite{rankingforabductive} proposed to reformulate the $\alpha$NLI task as a ranking problem. They use a 
learning-to-rank framework that contains a scoring function
and a loss function. 

\section{Experimental Results}
\begin{table}[t!]
\small
\centering
\scalebox{0.90}{{
\begin{tabular}{@{}l@{~}c@{~~~~}c@{~}}
\toprule
{\bf Model} & {\bf Dev (\%)} & {\bf Test (\%)}\\\midrule
Majority $^\diamond$ &{50.8}&{--}\\
GPT $^\diamond$ & {62.7} & {62.3} \\
BERT -L $^\diamond$ &{69.1}& {68.9} \\
McQueen \cite{mitra2019exploring} & {86.68} & {84.18}\\
\rowcolor{gray!25}
\textbf{Concurrent Work} &&\\
$L2R^2$ \cite{rankingforabductive}&{--} & {86.81}\\
\rowcolor{gray!25}
\textbf{This work}&{}&{}\\
Transformer Enc.\ w/o {LM$-$Pretraining} & {52.12} & {51.25}\\
 + MHKA & {54.96} & {53.91}\\
\hline
RoBERTa$-$B &{71.2}$\pm{0.3}$&{71.13}$\pm{0.5}$ \\
\rowcolor{blue!15}
RoBERTa$-$B + MHKA &{73.87}$\pm{0.2}$& \textbf{74.17}$\pm{0.2}$\\
RoBERTa$-$L & {85.06}$\pm{0.7}$ &{84.48}$\pm{0.7}$ \\
\hline
RoBERTa$-$L + Joint Training & {85.58}$\pm{0.5}$ & {84.91}$\pm{0.7}$\\
\rowcolor{blue!15}
RoBERTa$-$L + MHKA &{87.44}$\pm{0.5}$ & \textbf{87.12}$\pm{0.5}$ \\
\hline 

\textit{Human Perf.} &--& {91.4}\\
\bottomrule
\end{tabular}}}
\caption{Results on $\alpha$NLI dataset, $^\diamond$: as in \citet{bhagavatula2019abductive}, 
L = Large, B = Base, excluding unpublished leaderboard submissions}
\label{tab:results}
\end{table}

This section describes the experiments and results of our proposed model in different configurations. 

\textbf{Results on $\alpha$NLI.} Our experiment results for the \textbf{$\alpha$NLI task} are summarized in Table \ref{tab:results}. 
We compare performances of the following models: majority baseline, pre-trained LM baselines, and MHKA fine-tuned on RoBERTa-B(ase)/-L(arge). 
We observe consistent improvements
of our MHKA method over RoBERTa-B ($+3.04$ percentage points, pp.) and RoBERTa-L ($+2.64$ pp.) on $\alpha$NLI. Since MHKA 
uses RoBERTa to encode the input, this gain is mainly attributed to the use of knowledge and the multi-head knowledge attention technique. 
To better understand the impact of knowledge from pre-trained LMs, we trained a transformer encoder model 
\textit{without} fine-tuning on a pretrained LM (see Table \ref{tab:results}). 
Clearly, the overall performance of such a model drops considerably compared to the SOTA supervised models, but the improvement of MHKA 
by $+2.84$ points 
suggest that the impact of knowledge and reasoning obtained through multi-head knowledge attention is stable and independent from
the power of LMs.
Further, we compare our knowledge incorporation technique with \textit{Joint Training}: 
this method 
uses pre-trained LMs to jointly encode both task-specific input and the knowledge ([CLS] (K)nowledge [SEP] (I)nput text). 
Table \ref{tab:results} shows that \textit{Joint Training} yields limited improvement (+0.43 pp.) over the 
RoBERTa-L baseline -- the intuitive reason being that the pretrained LMs were never trained on such structured knowledge.\footnote{They also have a disadvantage when the length of context $+$ knowledge increases, as this causes a bottleneck for computation on a GPU with limited memory (8-24GB).} 
However, 
our \textit{MHKA model}
shows a solid improvement of $2.64$ pp.\ over the baseline. 
This suggests the impact of the \textit{Multi-Head Knowledge Attention} integration technique.

\begin{figure}[t]
  \centering
    \includegraphics[scale=1.0,height=4cm, width=0.3\paperwidth]{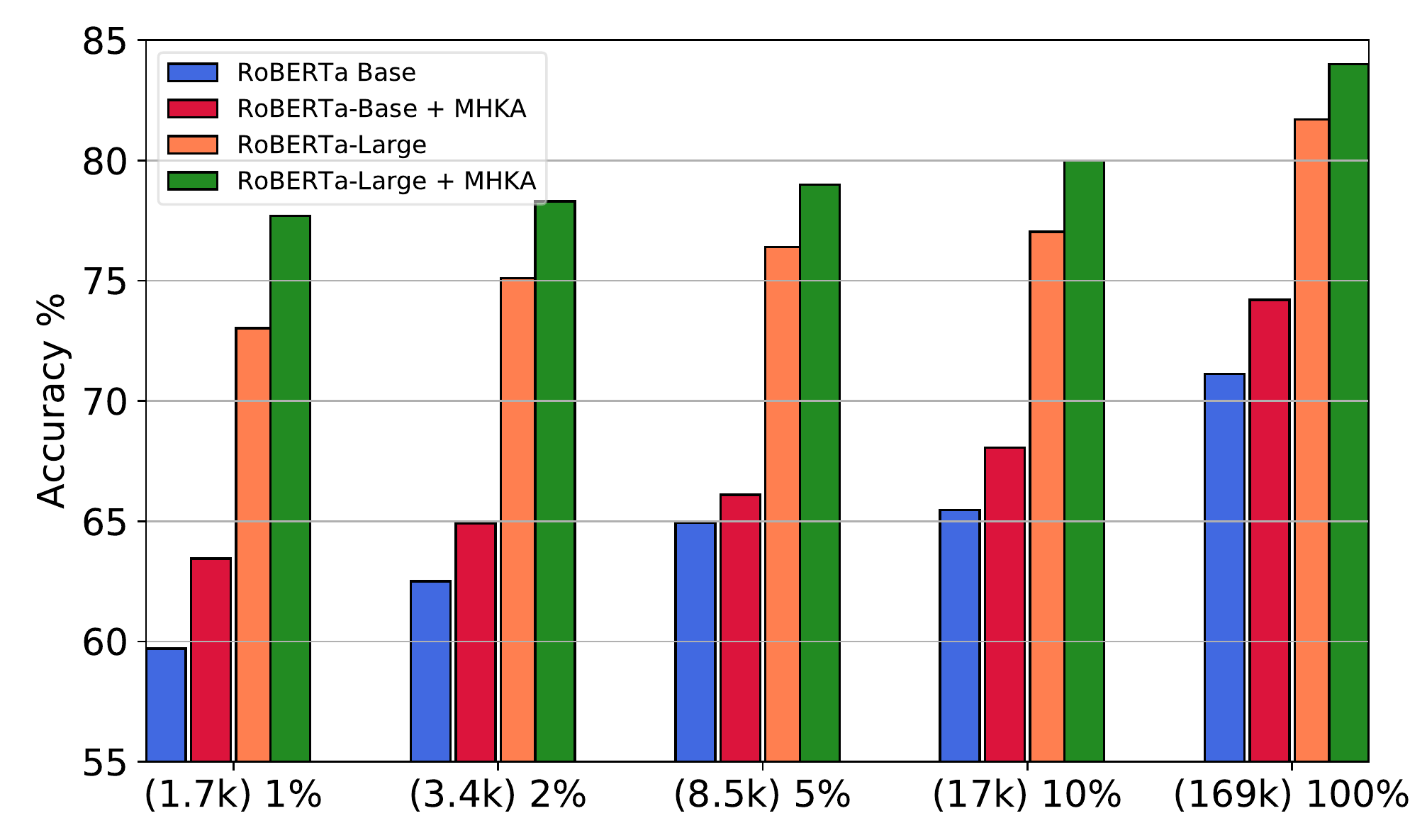}
    \caption{Accuracy for $\alpha$NLI  (Low Resource Setting)}
    \label{fig:low}
\end{figure}

\textbf{Low Resource Setting for $\alpha$NLI.} To better understand the impact of dataset scale on the performance of MHKA,
and to test its robustness  to data sparsity on $\alpha$NLI, we investigate low-resource scenarios where we
only use ${\{1, 2, 5, 10, 100\}\%}$ of the training data.  Figure \ref{fig:low} shows constant advances of MHKA over both RoBERTa-Base and -Large. 
This result indicates
the importance of knowledge in low-resource settings.
\begin{table}[t!]
\small
\centering
{
\scalebox{0.87}{
\begin{tabular}{@{}llll@{}}
\toprule
{\bf Model} & \bf{Input format}& {\bf Dev\%}& {\bf Test\%}\\
\midrule
RoBERTa-B & {$s_1$, [SEP], $s_2'$, [SEP], $s_3$}& {63.29} & {61.8}\\
 & {$s_1$, $s_2$ [SEP] $s_1$, $s_2'$}& {57.44} & {58.9}\\
 & {$s_1$, $s_2$, $s_3$ [SEP] $s_1$, $s_2'$ }& {64.38} & {62.8}\\
 & {$s_1$, $s_2$, $s_3$ [SEP] $s_1$, $s_2'$ , $s_3$}& {66.66} & {67.98}$\pm{0.5}$\\
 \rowcolor{blue!15}
 { }+ MHKA & {$s_1$, $s_2$, $s_3$ [SEP] $s_1$, $s_2'$ , $s_3$}& {69.34} & {69.7}$\pm{0.6}$\\
 RoBERTa-L & {$s_1$, $s_2$, $s_3$ [SEP] $s_1$, $s_2'$ , $s_3$}& {72.4} & {71.95}$\pm{0.6}$\\
 \rowcolor{blue!15}
 {  }+ MHKA & {$s_1$, $s_2$, $s_3$ [SEP] $s_1$, $s_2'$ , $s_3$} & {74.4}&{73.05}$\pm{0.3}$ \\\hline
\textit{Human Perf.}& & {}&\textbf{84.8}\\
\bottomrule
\end{tabular}}}
\caption{Results on Counterfactual Invariance Prediction (CIP). 
}\label{tab:results_counterfact}
\end{table}

\begin{table}[t!]
\small
\centering
{
\scalebox{1.0}{
\begin{tabular}{@{}lll@{}}
\toprule
{\bf Model} & {\bf Dev} & {\bf Test}\\
\midrule
RoBERTa-Large-$\alpha$NLI  & {76.3} &  {76.8}\\
Transfer Learning  & {78.00} & \textbf{79.04} \\
Transfer Learning + MHKA  & {78.6} & \textbf{80.77} \\
\bottomrule
\end{tabular}}
\caption{Impact of Counterfactual Invariance Prediction 
on $\alpha$NLI. Training data size for $\alpha$NLI is 8.5k ($5\%$)}\label{tab:mixed}}
\end{table}
\textbf{Results on CIP.} Table \ref{tab:results_counterfact} reports the results of our MHKA model on the CIP task, comparing to both RoBERTa baselines. 
 As this is a new task, we also report the results of RoBERTa-Base with different input formats. We find that providing the model with the full sequence ($s_1$, $s_2$, $s_3$ [SEP] $s_1$, $s_2'$ , $s_3$) gives best performance. 
By extending RoBERTa-Base and -Large with our MHKA reasoning component, we obtain an improvement of $+1.7$ and ${+1.1}$ percentage points, respectively.


%

\begin{figure}[t]
  \centering
    \includegraphics[scale=1.0,height=4.5cm]{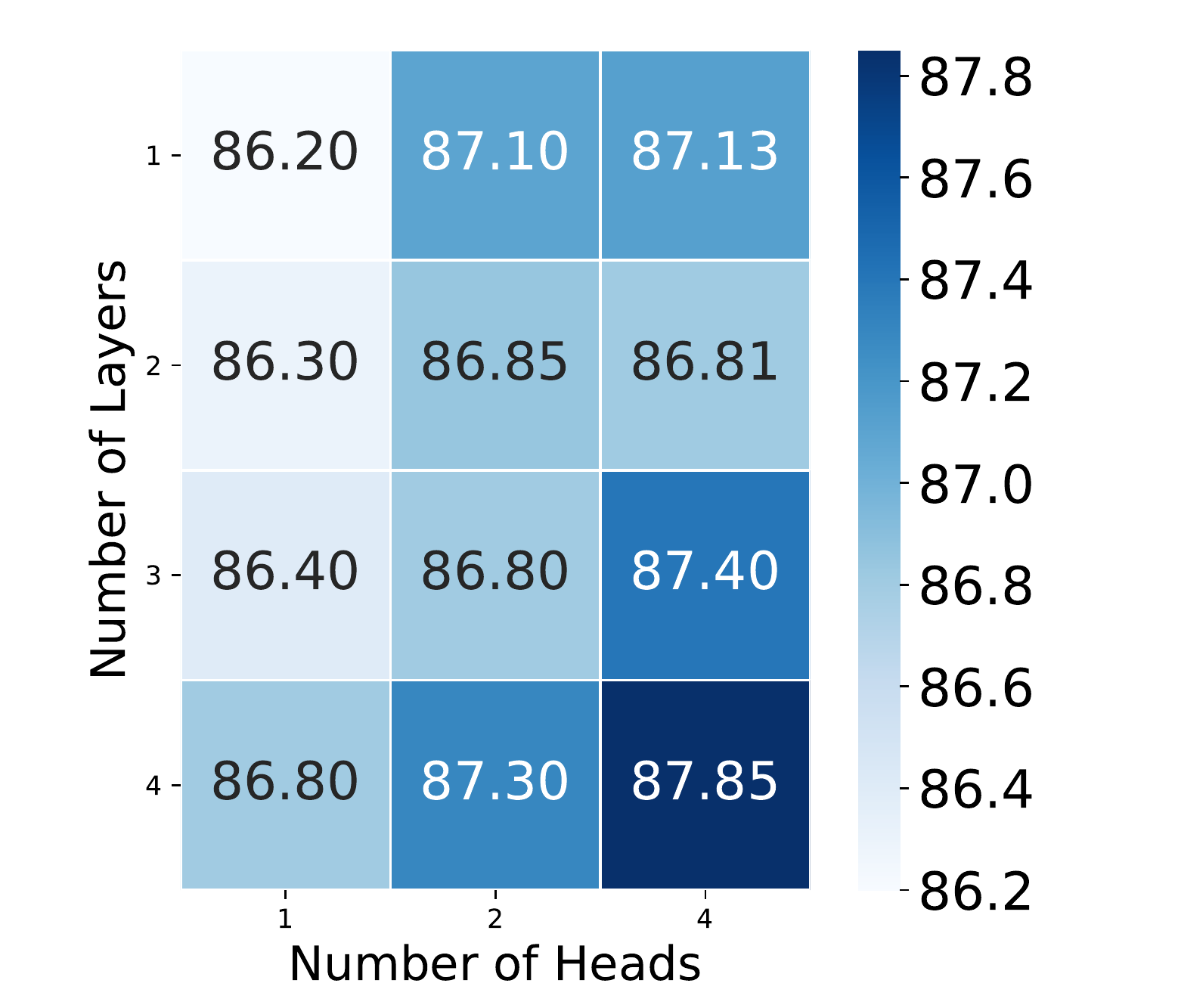}
    \caption{(a) Performance of MHKA model with different numbers of Heads and numbers of Layers.}
    \label{fig:ablation}
\end{figure}
\textbf{CIP for Transfer Learning.} We now test our hypothesis, discussed in \S \ref{sec:sec_mhka}, that a model trained on the CIP task can support the $\alpha$NLI task. 
We first fine-tune two models: RoBERTa-L and the RoBERTa-L+MHKA model on the CIP task (using the hyperparameters for the CIP task, Table \ref{tab:results_counterfact}). As a transfer-learning method, we 
fine-tune these models on $5\%$ of the training data for the $\alpha$NLI task (using the hyperparameters for $\alpha$NLI, Table \ref{tab:results}) and report the results in Table \ref{tab:mixed} as ``Transfer Learning" and ``Transfer Learning + MHKA". Table \ref{tab:mixed} also reports the results for RoBERTa-L trained on $5\%$ of the data of $\alpha$NLI (called RoBERTa-L-$\alpha$NLI).\footnote{The training data size of $\alpha$NLI is 14x larger than CIP. Therefore, in order to study the impact of CIP on $\alpha$NLI, we made the training data size of CIP and $\alpha$NLI comparable.} We obtain a +2.84 pp.\ improvement over this baseline by applying the pre-trained CIP model on the $\alpha$NLI task, and observe a further +1.73 pp.\ improvement (i.e., overall 3.97 points wrt.\ the baseline) with the stronger MHKA model.
These results confirm our hypothesis, and show that learning to distinguish the outcomes of factual and counterfactual events can help the model to better perform abduction. 



\if false 
\begin{table}[t!]
\small
\centering
{
\scalebox{1.0}{
\begin{tabular}{@{}l@{~}|l@{~}}
\toprule
{\bf Methods} & {\bf Accuracy} \\\midrule
\debjit{Transformer Encoder} w/o \af{LM-Pretraining} & {52.12}\\
 + MHKA & {54.96}  \\
\textbf{Pretrained Model} \\
RoBERTa (large) &{85.16}\\
+ Joint Training & {86.03}  \\
+ Attention-Layer & {86.21}  \\
+ MHKA & \textbf{87.85} \\
\bottomrule
\end{tabular}}}
\caption{Ablation results on $\alpha$NLI dataset for different Knowledge Integration Methods (on dev set)}\label{tab:ab_results}
\end{table}
\fi 


\textbf{Ablation on Reasoning Cell.} To give further insight into the factors for the model's capacity, we 
study the impact of the number of heads and layers in the 
reasoning cell. The left part of Figure \ref{fig:ablation}(a) shows the performance of the MHKA model with different numbers of heads and layers.
Note that the hidden dimensions of RoBERTa-Large is $1024$ which is not divisible by 3, therefore we have $1$, $2$, and $4$ as our attention heads. We observe that increasing the number of heads and layers improves the performance of the model. The intuitive explanation is that \textit{multiple heads} help the model to focus on multiple knowledge rules and at the same time \textit{multiple layers} help the model to recursively select the relevant knowledge rules. 


\section{Analysis}
\begin{figure}[t]
  \centering
    \includegraphics[scale=1.0,height=3cm]{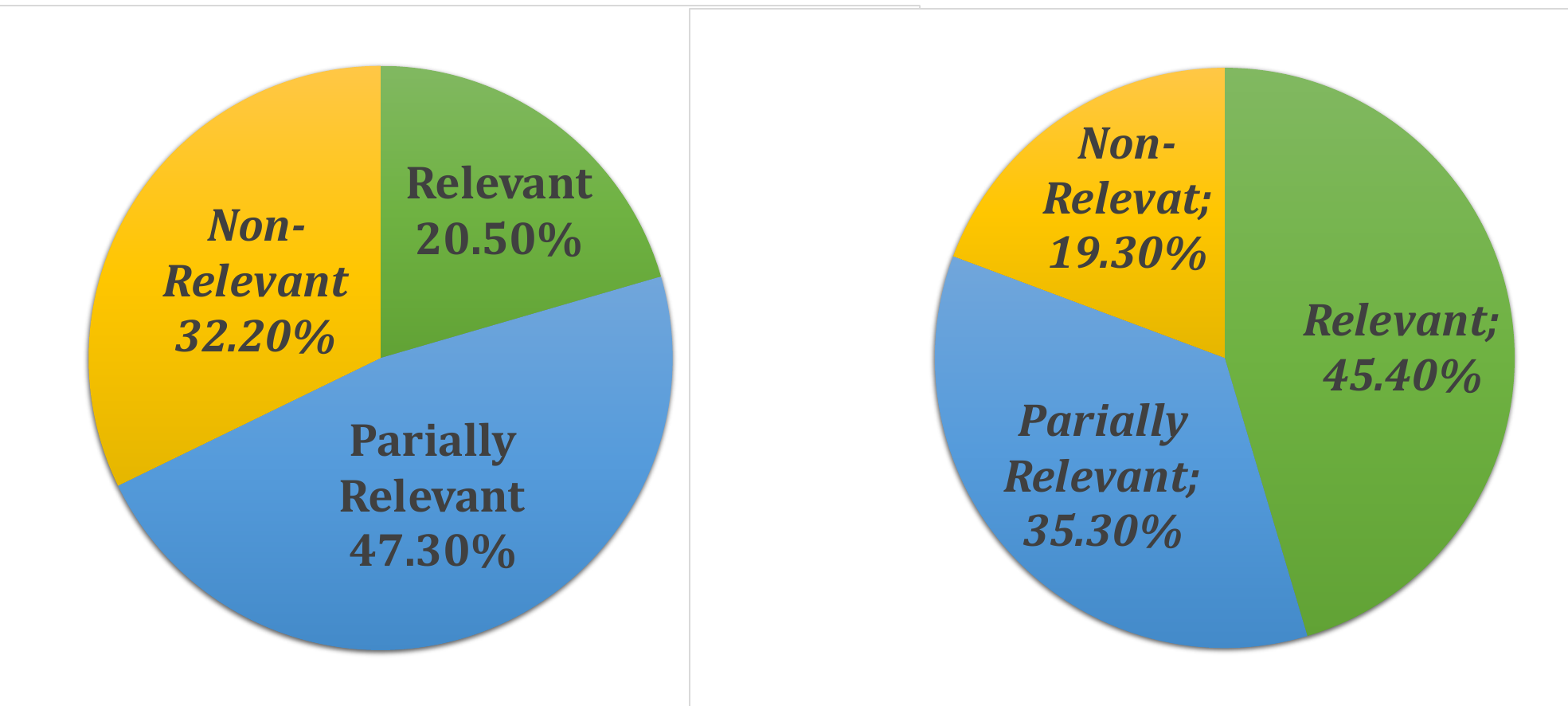}
    \caption{Human evaluation of the relevance of Knowledge Rules 
    a) for 100 instances from the $\alpha$NLI dev set and  b) 
    for the 56 (out of the 100) instances where the MHKA 
    model 
    predicted the correct hypothesis.}
    \label{fig:pie}
\end{figure}
\begin{table}[t]
    \centering
    \scalebox{0.78}{
    \begin{tabular}{@{}l@{~~}l@{~}l@{~}l@{~~}l@{}}
         \midrule
         &\bf{all know-} & \bf{w/o} & \bf{w/o relevant}  & \bf{replacing} \\ 
         &\bf{ledge}&\bf{irrelevant }& \bf{+ partially relevant} & \bf{relevant}\\\hline
         acc&56.2 & 57.6 (+1.4)& 49.4 ($-$6.8)& 45.05 ($-$11.2)\\
         \hline
         \# &56 & 54 ($-$2) & 20 ($-$36)& 18 ($-38$)\\
         \bottomrule
    \end{tabular}}
    \caption{\textit{row 1}: accuracy on 100 random instances from  $\alpha$NLI devset where the RoBERTa-L baseline fails; \textit{row 2}: nb.\ of instances (\#) correctly predicted by MHKA.} 
    \label{tab:removing_knowledge}
\end{table}

Up to now, we have focused on performance analysis with different experimental settings and model ablations to analyze our model's capacities. Now, we turn to leveraging the fact that our model works with semi-structured knowledge in order to obtain deeper insight into its inner workings.

\subsection{Quantitative Analysis.}
\textbf{Analysis on Knowledge Relevance.} We conduct human evaluation to validate the effectiveness and relevance of the extracted social commonsense knowledge rules. We randomly select 100 instances from the $\alpha$NLI dev set for which the RoBERTa-Large Baseline had failed
, along with their gold labels and the extracted knowledge. Table \ref{tab:removing_knowledge} 
shows that MHKA correctly predicts 56 instances correctly.
We asked two annotators to mark the knowledge rules that are relevant or partially relevant or irrelevant for each all 100 instances.
The obtained answers 
yield that in $20.50\%$ of cases the knowledge rules were relevant, in $47.30\%$ of cases they were partially relevant (see Figure \ref{fig:pie}.a). Figure \ref{fig:pie}.b depicts the relevance of knowledge rules for instances that are \textit{correctly predicted} by MHKA. 
The inter-annotator agreement had a Fleiss’ $\kappa$=0.62. 

\textbf{Analysis of Model's Robustness.} We then test the robustness of the models’ performance by manipulating the knowledge it receives for these instances in different ways: (a) we remove \textit{irrelevant} and (b) \textit{relevant} knowledge rules, (c) 
we manually change randomly selected rules from those that were found to be relevant by our annotators, and perturb them with artifacts. E.g., where annotators found that  “PersonX’s feelings” is relevant, we change the sentiment by choosing incorrect possible values from ATOMIC; for other relation types, we replace COMET’s generated object with an antonym “PersonX wanted to be [nice $\rightarrow$ mean]”, etc.
We evaluate the effect of the perturbations i) on all 100 instances, and ii) on the 56 correctly predicted instances. Results are shown in
Table \ref{tab:removing_knowledge}. 
We see, for (a),
a small improvement over the model results when using all knowledge, whereas for (b) and (c) an important performance drop 
occurs.  
For the 56 instances that MHKA resolves
correctly, 
for (b) and (c) we find the same effect, but with a much more drastic drop in performance for (b) and (c).  

This suggests that when the model is provided with relevant knowledge rules, it is able to utilize the knowledge well to 
perform the inference task. 
\begin{table}[t]
    \centering
    \scalebox{0.8}{
    \begin{tabular}{l|l|l}
         \midrule
         \bf{All} & \bf{Removing relevant} & \bf{Removing relation} \\
         \bf{knowledge}& \bf{relation tuples} & \bf{tuples randomly} \\\hline
         87.85 & 85.4 ($-$2.45)& 86.9 ($-$0.95)\\
         \bottomrule
    \end{tabular}}
    \caption{Accuracy on $\alpha$NLI (dev set)}
    \label{tab:probing}
\end{table}

In another test, we
remove knowledge rules with relations which were found most relevant by our annotators (namely, \textit{`PersonX's intent', `PersonX's want', `PersonX's need', `effect on PersonX', `effect on other', `PersonX feels'})
(see \textit{Supplement} for details). 
Table \ref{tab:probing} reports the results on dev set. 

We observe:  (a) when we remove the \textit{relevant} relational knowledge rules, the accuracy drops by 2.4 pp.\ suggesting that the model is benefitting from the knowledge rules. (b) when we remove 
knowledge rules \textit{randomly}, the accuracy drop is minimal which shows the robustness of our model.

\begin{figure}[t]
  \centering
    \includegraphics[scale=1.0,height=6cm, width=0.38\paperwidth]{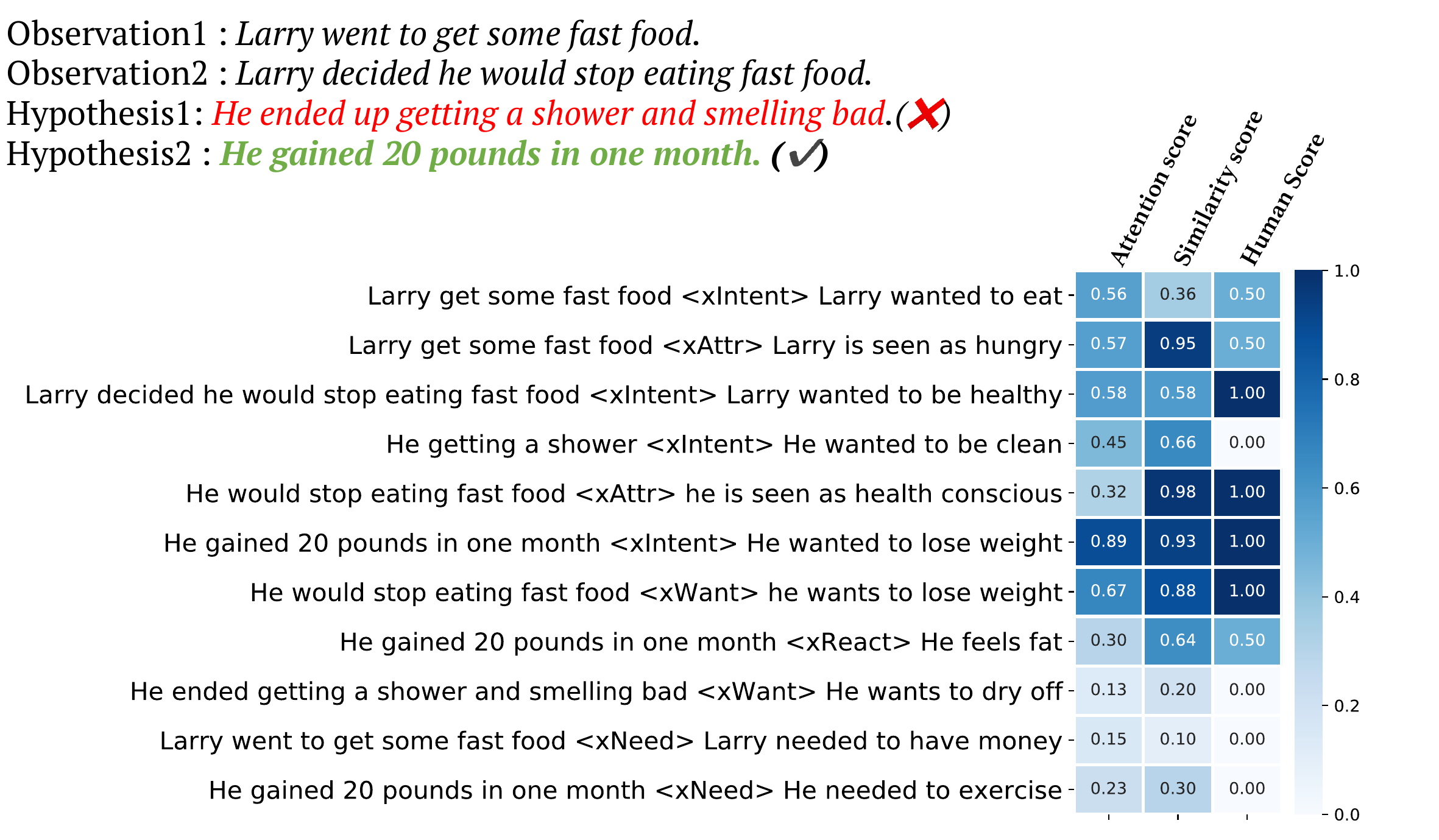}
    \caption{Comparing relevance scores of knowledge.}
    \label{fig:attention}
\end{figure}

\subsection{Qualitative Analysis.} 
Finally, we perform a study to better understand which knowledge rules were \textit{``used or incorporated in the Reasoning Cell}'' during the inference.

\textbf{A case study.} Figure \ref{fig:attention} depicts an example from the \textit{$\alpha$NLI} task where we see the context at the top, and knowledge rules along with different scores below. The \textit{Human scores} are annotated by the annotators where, $1.0$ = Relevant, $0.50$ = Partially relevant, $0.0$ = Irrelevant. We also show
the normalized attention scores over the structured knowledge rules\footnote{Note that we do not consider the attention maps as explanations. We assume that attention exhibits an intuitive interpretation of the model's inner workings.}. 
We also measure a similarity score (using dot product) between the final representation of the Reasoning cell and different knowledge rules. Intuitively, we expect that relevant knowledge rules should be incorporated in the final representation of the Reasoning cell, and therefore, should have a higher similarity score compared to irrelevant knowledge rules. Figure \ref{fig:attention}, illustrates one such example where we see that some relevant knowledge (judged by annotators) -- \textit{``He gained 20 pounds in one month $\langle$xIntent$\rangle$ He wanted to lose weight"}, and \textit{``He would stop eating fast food $\langle$xWant$\rangle$ he wants to lose weight''} -- are highly attended, and scored higher in similarity measure compared to others, indicating that the Reasoning Cell incorporated these knowledge rules. 
To study this further, we randomly selected $10$ instances from the \textit{$\alpha$NLI} dev set along with the knowledge rules. We found for $7$ out of $10$ instances that the MHKA model gave higher similarity scores to relevant or partially relevant knowledge rules than to irrelevant ones. 
\if false
\def\checkmark{\tikz\fill[scale=0.4](0,.35) -- (.25,0) -- (1,.7) -- (.25,.15) -- cycle;} 
\begin{table}[t!]
\small
\centering
\scalebox{0.5}{
\begin{tabular}{@{}lll@{}}
\toprule
{\bf Knowledge Rules} & {\bf Human} & {\bf Similarity}\\
\midrule
Larry get some fast food $\langle$xIntent$\rangle$  Larry wanted to eat & Partially Relevant & {} \\
Larry get some fast food $\langle$xAttr$\rangle$ Larry is seen as hungry &Irrelevant& {} \\
Larry decided he would stop eating fast food $\langle$xIntent$\rangle$ Larry wanted to be healthy &Relevant& {}\\
He getting a shower $\langle$xIntent$\rangle$ He wanted to be clean & Irrelevant & {} \\
He gained 20 pounds in one month $\langle$xIntent$\rangle$ He wanted to lose weight & Relevant & {} \\
He would stop eating fast food $\langle$xAttr$\rangle$ he is seen as health conscious & Relevant & {} \\
He would stop eating fast food $\langle$xWant$\rangle$ he wants to lose weight & Relevant & {} \\
He gained 20 pounds in one month $\langle$xReact$\rangle$ He feels fat & Relevant & {} \\
He ended getting a shower and smelling bad $\langle$xWant$\rangle$ He wants to dry off & Irrelevant & {} 
\\\bottomrule
\end{tabular}}
\caption{Model ablations for Reiss Classification on \textit{MNPCSCS} dataset w/o \textit{belonging}.}\label{tab:model_ab}
\end{table}
\fi



\if false
\begin{figure}
\begin{subfigure}{.5\textwidth}
  \centering
  \includegraphics[width=.2\linewidth]{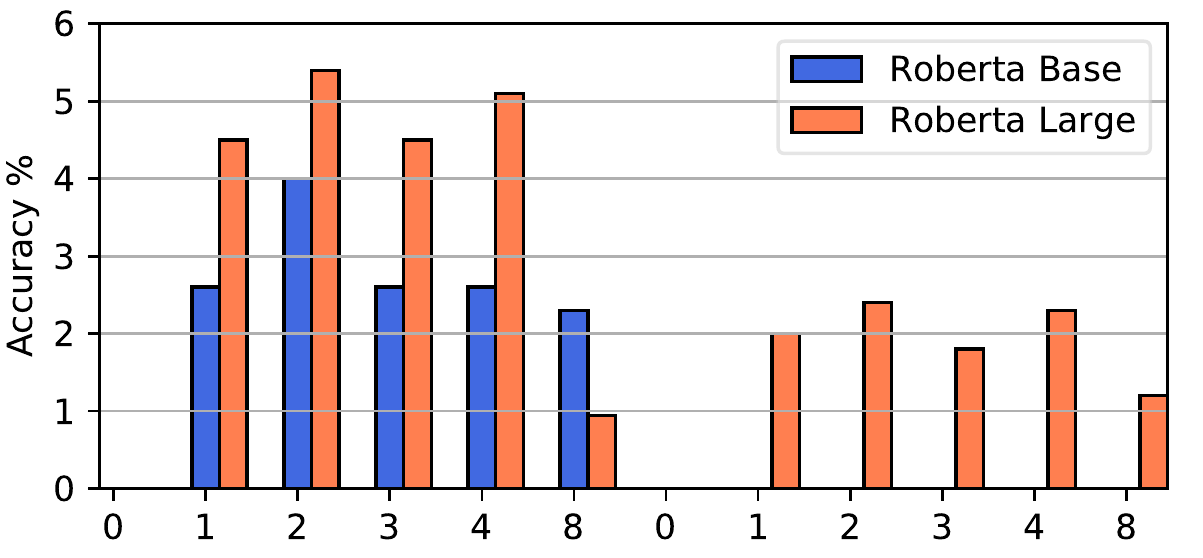}  
  \caption{Put your sub-caption here}
  \label{fig:sub-first}
\end{subfigure}
\begin{subfigure}{.5\textwidth}
  \centering
  \includegraphics[width=.2\linewidth]{emnlp2020-templates/figure/head.pdf}  
  \caption{Put your sub-caption here}
  \label{fig:sub-second}
\end{subfigure}
\newline
\begin{subfigure}{.5\textwidth}
  \centering
  \includegraphics[width=.2\linewidth]{emnlp2020-templates/figure/head.pdf}  
  \caption{Put your sub-caption here}
  \label{fig:sub-third}
\end{subfigure}
\begin{subfigure}{.5\textwidth}
  \centering
  \includegraphics[width=.2\linewidth]{emnlp2020-templates/figure/head.pdf}  
  \caption{Put your sub-caption here}
  \label{fig:sub-fourth}
\end{subfigure}
\caption{Put your caption here}
\label{fig:fig}
\end{figure}
\fi

\if false
\textbf{Does the Depth in Reasoning Cell resembles the number of the relations required?}\\
\begin{figure}[t]
  \centering
    \includegraphics[scale=1.0,height=3.5cm]{emnlp2020-templates/figure/head.pdf}
    \caption{Number of Layers in Reasoning Cell}
    \label{fig:example}
\end{figure} 

\textbf{Performance on Different Reasoning Types}\\
\begin{center}
  \begin{table}[!tbp]
      \scalebox{0.8}{
      \centering
      \begin{tabular}{@{}p{30mm}|p{50mm}}
          \hline
            \textbf{Dataset} & \textbf{Reasoning Types} \\ \hline
            SocialIQA & wants, motivations, effects, reactions, needs\\\hline
            Abductive Inference & wants, motivations, effects, ordering, negation, spatial, world knowledge, emotion\\ \hline
            Counterfactual Reasoning & effects, wants, needs, motivations, negation, situational fact, emotion \\\hline
      \end{tabular}}
      \caption{Different types of inferential reasoning for each dataset}
      \label{tab:reasoning}
  \end{table}
\end{center}

\begin{table}[t!]
\small
\centering
{
\scalebox{1.0}{
\begin{tabular}{@{}l@{~}|l@{~}}
\toprule
{\bf Methods} & {\bf AbductiveNLI} & {\bf SocialIQA*} \\\midrule
w/o Pretrained & {52.12} & {46.64} \\
 + MHKA & {54.96} & {48.94} \\
\textbf{Pretrained Model} \\
RoBERTa (large) &{82.02} &{73.1}\\
+ Joint & {82.57} &{73.36} \\
+ Attention-Layer & {82.6} & {74.3} \\
+ MHKA &{84.44} & {76.2} \\
\bottomrule
\end{tabular}}}
\caption{Ablations results for different Knowledge Integration Methods (on dev set)}\label{tab:ab_results}
\end{table}
\fi

\section{Related Work}
\textbf{Social Commonsense Knowledge} Teaching machines to reason about daily events with commonsense knowledge has been an important component for natural language understanding \cite{pcs1959, Davis2015CommonsenseRA, Storks2019CommonsenseRF}. 
Given the growth of interest among researchers in commonsense reasoning, a large body of work has been focused on learning commonsense knowledge representations \cite{lenat1995cyc, espinosa2005eventnet, speer2017conceptnet, tandon2017webchild}. 
In this work, we address social commonsense reasoning, where knowledge about events and its implications is crucial.
\citet{rashkin-etal-2018-event2mind} (Event2Mind) proposed a
knowledge 
resource for
commonsense inference about people’s intentions and reactions in
everyday events. Later, \citet{sapatomic} (ATOMIC) extended the Event2Mind resource with substantially more events, and with nine dimensions (\textit{If-then} relation types) per event. There has also been work on automatically acquiring commonsense knowledge \cite{Li2016CommonsenseKB, bosselut-etal-2019-comet, malaviya2019exploiting}. Recently, \citet{glucose2020}  introduced a large-scale dataset (GLUCOSE) capturing ten dimensions of causal explanation (implicit commonsense knowledge) in a narrative context. However, learning to reason over such event-based semi-structured knowledge is still a challenging task. In this work, we propose a model which learns to imitate reasoning using such structured knowledge. 

\textbf{Commonsense Reasoning (CR):} There is a large body of research on commonsense reasoning over natural language text \cite{levesque2012winograd, bowman2015large, zellers2019hellaswag, trichelair-etal-2019-reasonable, beckeretal:2020b}. 
We
discuss the ones most related to our work. Earlier works sought to utilize 
rule-based reasoning or hand-crafted features \cite{sun1995robust, gupta2005commonsense}. With the increase in size of commonsense knowledge bases \cite{suchanek2007yago, speer2017conceptnet} researchers started utilizing them to help models perform commonsense reasoning \cite{schuller2014tackling, liu2017cause}. Recently, there have been attempts to leverage pre-trained language models to learn and perform commonsense inference, and they achieved state-of-the-art results
\cite{radford2018improving, trinh2018simple, kocijan2019surprisingly, radford2019language}. 
Our model takes advantage of both pre-trained LMs and structured knowledge, which allows us to inspect the reasoning process. We also demonstrate that our model shows strong performance for different, and finely structured tasks in abductive and counterfactual reasoning.



\textbf{Structured Commonsense Knowledge in Neural Systems:} Different approaches have been proposed to extract and integrate external knowledge into neural models for various NLU tasks such as reading comprehension (RC) \cite{xu2017incorporating, mihaylov-frank-2018-knowledgeable, weissenborn2017dynamic}, question answering (QA) \cite{xu-etal-2016-question, tandon-etal-2018-reasoning, wang2019improving}, etc. 
Recently, many works proposed different ways to extract knowledge from static knowledge graphs (KGs). Most notable are ones that
extract subgraphs from KGs using
either 
heuristic methods \cite{bauer2018commonsense} or graph-based ranking methods \cite{Paul2019RankingAS, pauletal:2020}, or else utilize know\-ledge graph embeddings \cite{lin2019kagnet} to rank and select relevant knowledge triples or paths. 

Similar to \citet{Bosselut2019DynamicKG} and \citet{Shwartz2020UnsupervisedCQ}, in this work we generate contextually relevant knowledge using language models trained on KGs.  
With the increase in performance of transformer-based models there has been a shift from RNN-based neural models to pre-trained LMs. 
Incorporating extracted knowledge using attention mechanism (single dot product) has become a standard procedure. However, we propose a multi-head attention model that can recursively select multiple generated structured knowledge rules, 
and also allows inspection by analyzing the used knowledge. 

\section{Conclusion}
In this work, we propose a new \textit{multi-head knowledge attention model} to incorporate semi-structured social commonsense knowledge.  
We show that our model improves over state-of-the-art LMs on two complex commonsense inference tasks. 
Besides the improvement i) we demonstrate a correlation between abduction and counterfactual reasoning in a narrative context, based on the newly proposed task of counterfactual invariance prediction, which we apply to support abductive inference.   Importantly, ii) we confirm the reasoning capacity of our model by perturbing and adding noise to the knowledge, and performing model inspection using manually validated knowledge rules. In future work, we aim to deeper investigate compositional effects of inferencing, such as the interaction of socially grounded and general inferential knowledge.


\subsection*{Acknowledgements}
This work has been supported by the German Research Foundation as part of the Research
Training Group “Adaptive Preparation of Information from Heterogeneous Sources” (AIPHES)
under grant No.\ GRK 1994/1.  
We thank our annotators for their valuable annotations. We also thank NVIDIA Corporation for donating GPUs used in this research.

\bibliography{emnlp2020}
\bibliographystyle{acl_natbib}
\appendix
\section{Supplement Material}

\subsection{Data}

\textbf{Counterfactual Invariance Prediction (CIP)} In this work, we define a new task that tests the capability of models to predict whether under the assumption of
a counterfactual event, a (later) factual event remains invariant or not in a narrative context. The formal setup is: given the first three consecutive sentences from a narrative story $s_1$ (premise), $s_2$ (initial context), $s_3$ (factual event) and an additional sentence ${s'}_2$ that is counterfactual to the initial context $s_2$, the task is to predict whether $s_3$ is invariant given $s_1$, ${s'}_2$ or not.  Table \ref{tab:supplementary_example}, display some examples of CIP task. 

\textbf{How was the data collected?} 
\cite{qin-counterfactual} proposed a dataset to encourage models to learn how to rewrite stories with counterfactual reasoning.  They build the dataset on top of the ROCStories corpus, which comprises of five-sentence stories $S$ = ($s_1$, $s_2$, ..., $s_5$).
The formal setup is: $s_1$ (premise), $s_2$ (initial context), the last three sentences $s_{3:5}$ are the original ending of story. For each story they  ask crowdworkers to write a counterfactual sentence to the initial context $s_2$ and also re-write the ending $s'_{3:5}$ according to the counterfactual sentence. We automatically collect counterfactual invariance examples by checking if (original) $s_3$ == (edited) $s'_3$ and similarly, if (original) $s_3$ != (edited) $s'_3$ non-invariant examples from their dataset to create a balanced dataset for our proposed CIP task.

\subsection{Hyperparameter}
In all models the Reasoning Cell and the Knowledge Encoder are both instantiated by  a Transformer with 4 attention heads and depth = $4$. For each task, we select the hyperparameters that yield best performance on the dev set. Specifically, we perform a grid search over the hyperparameter settings with a learning rate in \{$1e$-$5$, $2e$-$5$, $5e$-$6$\}, a batch size in $\{4, 8\}$, and a number of epochs in $\{3, 5, 10\}$. Training is performed using cross-entropy loss. For evaluation, we measure accuracy. We report performance on the test sets by averaging results along with the variance obtained for 5 different seeds. We use Adam Optimizer, and drop-out rate = 0.1. The best hyperparameter details are stated in Table \ref{tab:supplementary_hyperparameter}. We experimented on GPU size of 11GB and 24GB. 
\begin{table}[t]
    \centering
    \begin{tabular}{|l|l|l|l|}
         \bf{Datasets} & \bf{Learning Rate} & \bf{Epochs} & \bf{Batch}\\\hline
         {$\alpha$NLI} &  {5e-6} & {5} & {8} \\
         CIP &  {5e-6} & {5} & {8}\\\hline
    \end{tabular}
    \caption{Best Hyperparameter}
    \label{tab:supplementary_hyperparameter}
\end{table}

\begin{figure}[t]
  \centering
    \includegraphics[scale=1.0,height=4cm]{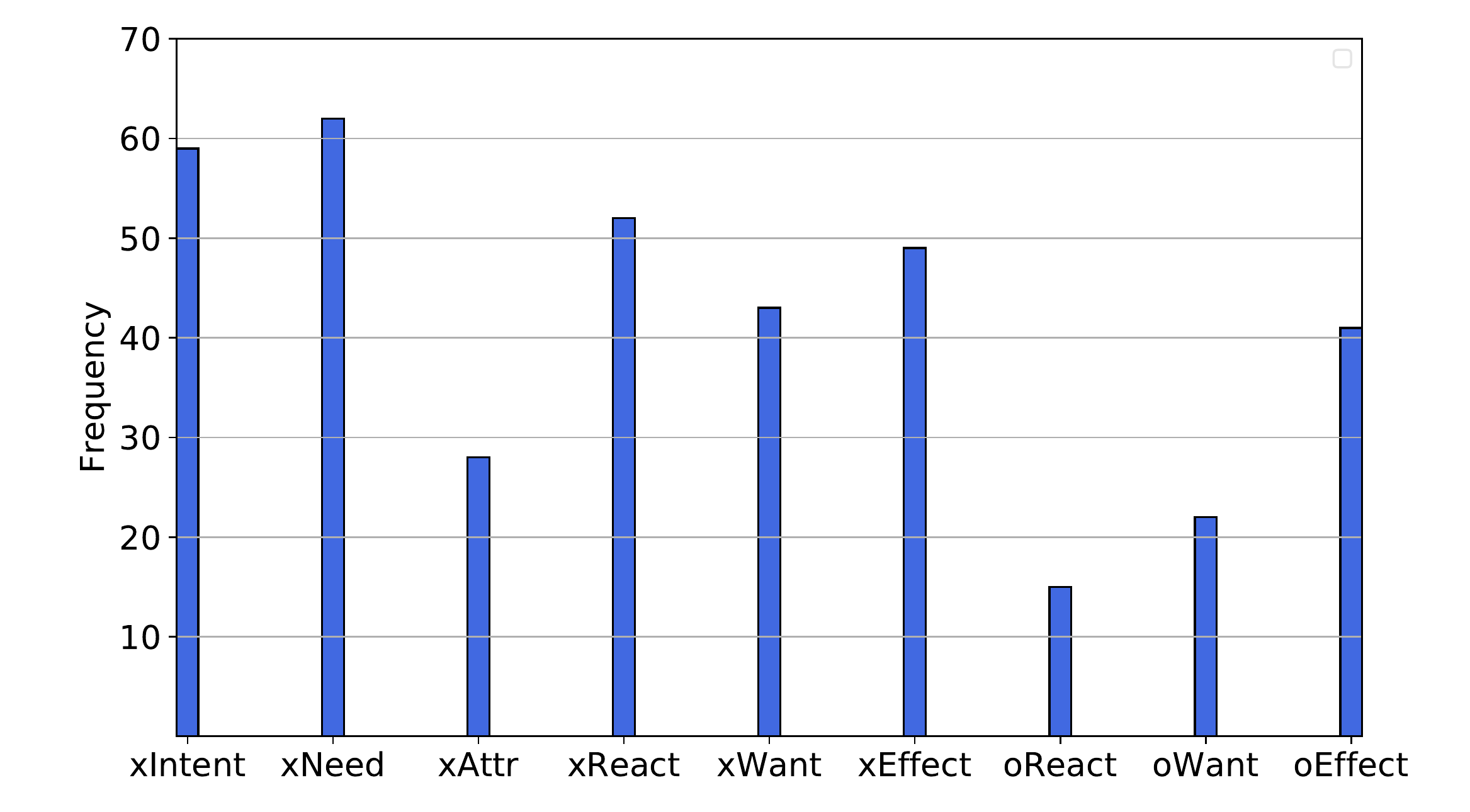}
    \caption{Distribution of relational Knowledge required per instances}
    \label{fig:supplementary_relational}
\end{figure}

\subsection{Human Analysis}
We conduct human evaluation to validate the effectiveness and relevance of the extracted social commonsense knowledge rules. We randomly selected 100 instances from the $\alpha$NLI dev set for which the RoBERTa-Large Baseline had failed, along with gold labels. Firstly, we asked two annotators to construct the knowledge rules required to  perform the $\alpha$NLI task for these 100 instances, using the different dimensions of ATOMIC knowledge.
We found that for ${90}$ instances such relational knowledge is required, and on average $3$ knowledge rules are important. This study suggests that models need to find a chain of rules to perform the inference task. Further, we test the robustness of the models’ performance by removing random knowledge rules vs. removing knowledge rules with relations which were found most relevant by our annotators (namely, 'PersonX intent', ‘PersonX’s want’, ‘PersonX’s need’,
‘effect on PersonX’, ‘effect on other’, ‘PersonX feels’) see Figure \ref{fig:supplementary_relational}.



\subsection{Attention Visualization}
We study the visualization the attention distributions over different (relation) dimensions of ATOMIC, produced by our MHKA (Reasoning Cell) model. It depicts the attention distribution and change in attention over multiple structured social knowledge rules (relations), and over different layers. It also allows to inspect inner working of the Reasoning cell. $\alpha$NLI Examples: \textbf{Observation1:} \textit{Dotty was being very grumpy.} \textbf{Observation2:} \textit{She felt much better afterwards.} \textbf{Hypothesis:} \textit{Dotty call some close friends to chat.} The model correctly attended the relation `xwant, xintent, xneed, effect on others'.

\begin{table}[!tbp]
      \scalebox{0.7}{
      \centering
      \begin{tabular}{@{}p{20mm}|p{60mm}|p{40mm}}
          \hline
            \textbf{Relation} & \textbf{Question} & \textbf{Textual Description} \\ \hline
            {xIntent} & {Why does X cause the event?} & {'because PersonX wanted'} \\
            {xNeed} & {What does X need to do before the event?} & {'PersonX needed'} \\
            {xAttr} & {How would X be described?} & {'PersonX is seen as'} \\
            {xReact} & {How does X feel after the event?} & {'PersonX feels'} \\
            {xWant} & {What would X likely want to do after the event?} & {'PersonX wants'} \\
            {xEffect} & {What effects does the event have on X?} & {'effect on PersonX'} \\
            {oReact} & {How do others' feel after the event?} & {others feel}\\
            {oWant} & {What would others likely want to do after the event} & {others wants' } \\
            {oEffect} & {What effects does the event have on others?} & {'effect on others'} \\
      \end{tabular}}
      \caption{The taxonomy of if-then reasoning types from ATOMIC \cite{sapatomic}. }
      \label{tab:supplementary_sapatomic}
\end{table}

\begin{center}
\begin{table*}[!tbp]
      \scalebox{0.8}{
      \centering
      \begin{tabular}{@{}p{145mm}|p{20mm}}
          \hline
            \textbf{Context \& Questions} & \textbf{Options} \\ \hline
            \textit{Bob had to get to work in the morning. His car battery was struggling to start the car. He called his neighbor for a jump start.} 
            \textbf{Alternatively:} \textit{Bob had to get to work in the morning. His car won't start.} & answer [Yes] \\\hline
            \textit{Bill and Teddy were at the bar together. Bill noticed a pretty girl. He went up to her to flirt.}
            \textbf{Alternatively:} \textit{Bill and Teddy were at the bar together. Bill noticed his mom was there.} & answer [No] \\\hline
            \textit{I loved to eat honey with my oatmeal. One day I unexpectedly ran out of honey.	I did not want to eat my oatmeal without honey.} 
            \textbf{Alternatively:} \textit{I loved to eat honey with my oatmeal. One day I realized that maple syrup was even better with my oatmeal.} & answer [No] \\\hline
            \textit{I went to las vegas. I learned that i really like the slot machines.	I spent a lot of time on them.}	
            \textbf{Alternatively:} {I went to las vegas. I learned that slot machines are a great way to make money.} & answer [Yes] \\\hline
            \hline
      \end{tabular}}
      \caption{CIP Examples}
      \label{tab:supplementary_example}
  \end{table*}
\end{center}


\begin{figure*}[t]
  \centering
    \includegraphics[scale=1.0,height=9cm]{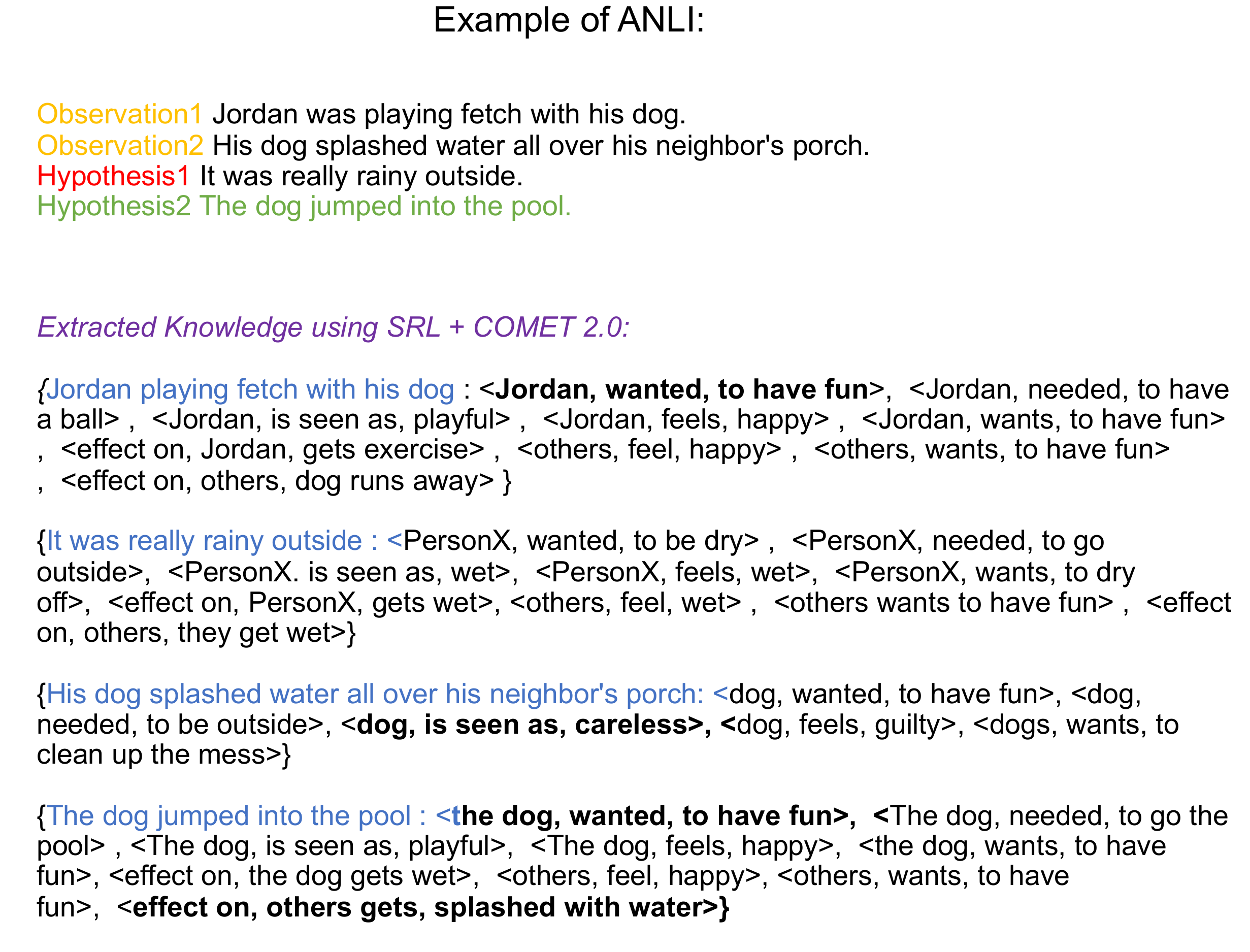}
    \caption{Example}
    \label{fig:supplementary_anli}
\end{figure*}


\end{document}